\documentclass{article}

\PassOptionsToPackage{numbers,sort&compress}{natbib}
\usepackage[final]{neurips_2026}

\usepackage[utf8]{inputenc}
\usepackage[T1]{fontenc}
\usepackage[colorlinks=true,linkcolor=blue,citecolor=blue,urlcolor=blue]{hyperref}
\usepackage{url}
\usepackage{booktabs}
\usepackage{tabularx}
\usepackage{enumitem}
\usepackage{amsfonts}
\usepackage{amsmath}
\usepackage{amssymb}
\usepackage{nicefrac}
\usepackage{microtype}
\usepackage{graphicx}
\usepackage{multirow}
\usepackage{algorithm}
\usepackage{algpseudocode}
\usepackage{wrapfig}
\usepackage{float}
\usepackage[table]{xcolor}

\usepackage{caption}
\newcommand{\Co}{\textsc{Co-PiLOT}}
\newcommand{\meridian}{\textsc{Meridian}}
\newcommand{\dante}{\textsc{Dante}}
\newcommand{\turbo}{\textsc{TuRBO}}
\newcommand{\baxus}{\textsc{BAxUS}}
\newcommand{\saasbo}{\textsc{SaasBO}}
\newcommand{\damask}{\textsc{Damask}}
\newcommand{\dreamthreed}{\textsc{Dream.3D}}

\newcommand{\Z}{\mathcal{Z}}
\newcommand{\X}{\mathcal{X}}
\newcommand{\Dset}{\mathcal{D}_t}
\newcommand{\Enc}{\mathcal{E}}
\newcommand{\Dec}{\mathcal{D}}

\newcommand{\Opt}{\mathcal{O}}
\newcommand{\R}{\mathbb{R}}
\newcommand{\E}{\mathbb{E}}
\newcommand{\norm}[1]{\left\lVert#1\right\rVert}

\newcommand{\sigy}{\sigma_y}
\newcommand{\sigu}{\sigma_u}

\newsavebox{\tabvtwobox}
\newlength{\figvtwowidth}

\newcommand{\urlEncDec}{\url{https://github.com/mahishguru/microstructure-encoder-decoder}}
\newcommand{\urlCodec}{\url{https://github.com/mahishguru/orientation-codec}}
\newcommand{\urlMeridian}{\url{https://github.com/mahishguru/meridian}}
\newcommand{\urlData}{\url{https://zenodo.org/records/23036836}}

\title{\Co{}: Constrained Physics-Informed Latent Optimization for Target-Driven Inverse Design}

\author{%
  Mahish K.\ Guru$^{1,2}$ \quad
  Mayank Nagar$^{1}$ \quad
  Ayush Vyas$^{1}$ \quad
  Jan Bohlen$^{1}$ \\
  \textbf{Roland Aydin}$^{3,4}$ \quad
  \textbf{Noomane Ben Khalifa}$^{1,2}$ \\[4pt]
  $^{1}$Institute of Material and Process Design, Helmholtz-Zentrum Hereon, Germany \\
  $^{2}$Institute of Production Technology and Systems, Leuphana University L\"uneburg, Germany \\
  $^{3}$AI for Physical Systems, German Research Center for Artificial Intelligence (DFKI), Germany \\
  $^{4}$Saarland University, Germany \\[2pt]
  \texttt{mahish.guru@hereon.de}
}
\makeatletter
\providecommand{\@trackname}{Preprint.}
\makeatother
\begin{document}

\maketitle

\begin{abstract}
Inverse design of physical systems (molecules, devices, microstructures) often reduces to optimizing a high-dimensional structure against an expensive black-box simulator. Direct search is difficult because the space is non-Euclidean, feasibility is hard to encode, and each evaluation is expensive. We present \Co{}, a latent optimization approach that maps candidates through a generative encoder--decoder, uses the decoder as a learned validity prior, and searches the latent space with physics-informed black-box optimization. The framework is applied on the inverse design of magnesium alloy microstructure/texture. We develop a vision transformer based--encoder; paired with latent diffusion, diffusion transformer and rectified-flow transformer--based decoders on $\sim80{,}000$ EBSD-derived microstructure dataset to learn a minimal bottleneck, $z$. The ViT-FMDiT model ($z$=$768$) reconstructs high-fidelity microstructure images (FID $27.86$, MS-SSIM $0.178$), which our self-segmenting orientation codec converts into input grids for crystal plasticity solver. Finally, we introduce \meridian{}, an active latent optimizer driven by deep-kernel Gaussian-process uncertainty, failure-aware feasibility prediction, manifold-aware trust regions, and target-aware acquisition. Within a budget of $160$ simulations, the ViT-FMDiT and \meridian{} combination yields the best target-driven objective score, reducing the relative target error by $3$--$22\%$ against seven baselines (\dante{}, \turbo{}, \baxus{}, CMA-ES, DDOM, SEIKO, DDPO) on the same decoder.
\end{abstract}

\section{Introduction}
\label{sec:intro}

Across science and engineering, the same template recurs: \emph{a designer is given a target behaviour and must find a physical object that realises it}. A medicinal chemist is given a binding-affinity profile and must find a molecule whose density-functional-theory (DFT) energetics match it~\citep{gomez2018, Yoo2023,Axelrod2022}; a photonics engineer is given a target far-field radiation pattern and must find a metasurface whose Maxwell-solver response matches it~\citep{Molesky2018,Zhou2021}; a crystallographer is given a desired band gap and must find a periodic crystal whose first-principles spectrum matches it~\citep{xie2022,zeni2025,jiao2023}. Inverse design asks: given a target property vector $p^\star$, find a design $x^\star \in \X$ such that $f(x^\star) \approx p^\star$, where $f$ is an expensive black-box oracle (e.g., a physical assay or simulator) and a feasibility constraint $g(x) \le 0$ encodes stability requirements. Three structural obstacles make direct search on $\X$ intractable. \emph{(i)}~$\X$ is discrete and non-Euclidean (graphs of atoms, fields of crystallographic orientations), so gradient methods do not apply through a simulator that has no usable adjoint~\citep{Liu2023}. \emph{(ii)}~most points in $\X$ are physically invalid, and feasibility cannot be expressed as a simple box, so unguided sampling almost never yields a candidate that oracle will run to convergence~\citep{Lu_2022}. \emph{(iii)}~Each evaluation $f(x)$ costs hours, with $10^2$--$10^3$ evaluation budgets that rules out brute-force sampling, GAN-inversion-by-backpropagation~\citep{creswell2018,li2026towards}, and high-dimensional Bayesian optimization (BO) over the raw space~\citep{eriksson2019,papenmeier2022, Frazier_2015}. Prior work side-steps the simulator by training a fast surrogate~\citep{yang2018,cang2018,xie2022,zeni2025}, trading simulator cost for surrogate-fidelity risk that breaks whenever the surrogate is asked to extrapolate past its training distribution.

To ground this framework, we tackle an industrial and scientific challenge that falls entirely out of the 'cheap-oracle' assumption dominating existing latent-optimization research: \emph{target-driven inverse design of polycrystalline Magnesium (Mg) alloy microstructures and textures}. The protagonist is the engineer with a target stress--strain curve, eg. a structural designer specifying a Mg-alloy AZ31 component for an automotive crash member~\citep{Abbott2016}; a clinician specifying a bioresorbable Mg--Gd bone screw whose yield strength must match cortical bone over a degradation window~\citep{Chiu}; an aerospace designer chasing an anisotropic high-toughness texture in a structural panel~\citep{Bai2023}. In every case the deliverable is not a microstructure \emph{image} but an arrangement of grains and crystallographic orientations that, when simulated through a crystal-plasticity (CP) oracle such as \damask{}~\citep{roters2019}, returns a property tuple $(\sigy, n, K, \sigu)$ within tolerance of the target. Microstructure inverse design is a hard problem because the design space comprises of grains with symmetric-crystallographic orientaions~\citep{Mason2009,Tuomo} and anisotropic, twin-dominated, and texture-coupled plasticity which causes visually similar microstructures to yield vastly different stress--strain responses~\citep{guru2025}; and a single $300{\times}300$ CP simulation takes $\sim\!6$ minutes and risks divergence~\citep{roters2019}. Inverse microstructure and texture design are therefore recognised as ill-conditioned, high-dimensional, and data-starved~\citep{generale2024,buzzy2025}.

\textbf{Why no off-the-shelf recipe applies?} To bypass the intractability of raw search space, one could theoretically map the problem into a continuous latent search space with a pretrained generator $\Dec : \R^d \to \X$. In such a setup, decoded samples lie near a learned manifold, the latent prior yields a free box constraint $\Z = [-B, B]^d$, and the optimization objective $J \circ f \circ \Dec$ is well-defined. This theoretically enables pure target-driven design, where optimal orientations and grain morphologies emerge directly from stress--strain requirements~\citep{Xiong2016}. \emph{In practice, however,} existing studies assume cheap oracles~\citep{tripp2020,gomez2018,maus2022,stanton2022}, high-dimensional BO lacks generative priors~\citep{eriksson2019,papenmeier2022,eriksson2021}, and recent crystal models~\citep{xie2022,zeni2025,jiao2023} are built for sampling, not inverse optimization. None can reliably navigate an expensive, occasionally divergent simulator while enforcing valid material-grain orientation fields.

\textbf{What is new.} The latent representation is an enabling component, not our contribution; the methodological claim concerns the optimizer. Existing active-learning optimizers assume that every query returns a valid value and search generic, isotropic regions. \meridian{} couples two mechanisms not jointly present in any of the seven baselines: \emph{failure-aware feasibility learning}, a classifier trained on failed decode/codec/solver evaluations that steers acquisition away from likely divergence, and \emph{manifold-aware local search}, a surrogate-shaped active-subspace trust region with a latent-shell projection that keeps candidates where the decoder yields valid microstructures. To our knowledge, \Co{} is also the first closed loop coupling an image-domain generative prior to a crystal-plasticity solver under a $160$-simulation budget.

\textbf{Contributions}. We present \Co{} (\textbf{Co}nstrained \textbf{P}hysics-\textbf{i}nformed \textbf{L}atent \textbf{O}ptimization for \textbf{T}arget-driven inverse design), a framework that lifts inverse design into a pretrained generative latent space and runs a constrained physics-informed black-box optimizer there.
\begin{enumerate}[leftmargin=1.2em,itemsep=-2pt,topsep=2pt]
  \item \textbf{General framework} (Sec.~\ref{sec:method}). We formulate inverse design as a 5-tuple $(\Dec, f, J, \mathcal{C}, \Opt)$ with independently swappable decoder, oracle, objective, constraint set, and optimizer. The abstraction is domain-agnostic; we demonstrate it on one domain (Mg-alloy microstructures) and leave other instantiations to future work.
    \item \textbf{Microstructure encoder--decoder family} (Sec.~\ref{sec:enc-dec}). A shared Vision Transformer (ViT) based encoder is paired with a primary Flow Matching Diffusion Transfomer (FMDiT) based decoder, trained jointly on $81{,}756$ EBSD-derived microstructures across 17 Mg-alloy classes.
    \item \textbf{Orientation codec} (Sec.~\ref{sec:codec}). An invertible map between RGB images and orientation fields with a closed-form decode and a grain-segmentation \dreamthreed{} writer that is used by the CP solver. It achieves ${\sim}0.7^\circ$ round-trip misorientation, below the $5^\circ$ grain-boundary threshold~\citep{readshockley1950, groeber2014}.
    \item \textbf{\meridian{}, a latent optimizer} (Sec.~\ref{sec:meridian}). We developed a combination of a deep-kernel Gaussian Process surrogate, a shared-trunk feasibility classifier, an active-subspace trust region with latent-shell projection, and a DPP batch selector; against seven baselines under a matched budget, ViT-FMDiT-$768$ with \meridian{} reaches the best target-driven and toughness-weighted objectives.
\end{enumerate}

The 5-tuple abstraction is domain-agnostic: in principle any setting with a pretrained decoder for $\X$ and a slow oracle scoring $\Dec(z)$ is in scope (molecules with a VAE and a DFT oracle~\citep{gomez2018}, photonic devices with an EM solver~\citep{Yeung2023}, topology optimization with an FEM oracle~\citep{nie2020}), but we demonstrate it only on Mg-alloy microstructures. All code is released (Sec.~\ref{sec:conclusion}).

\begin{figure}[t]
  \centering
  \includegraphics[width=\linewidth]{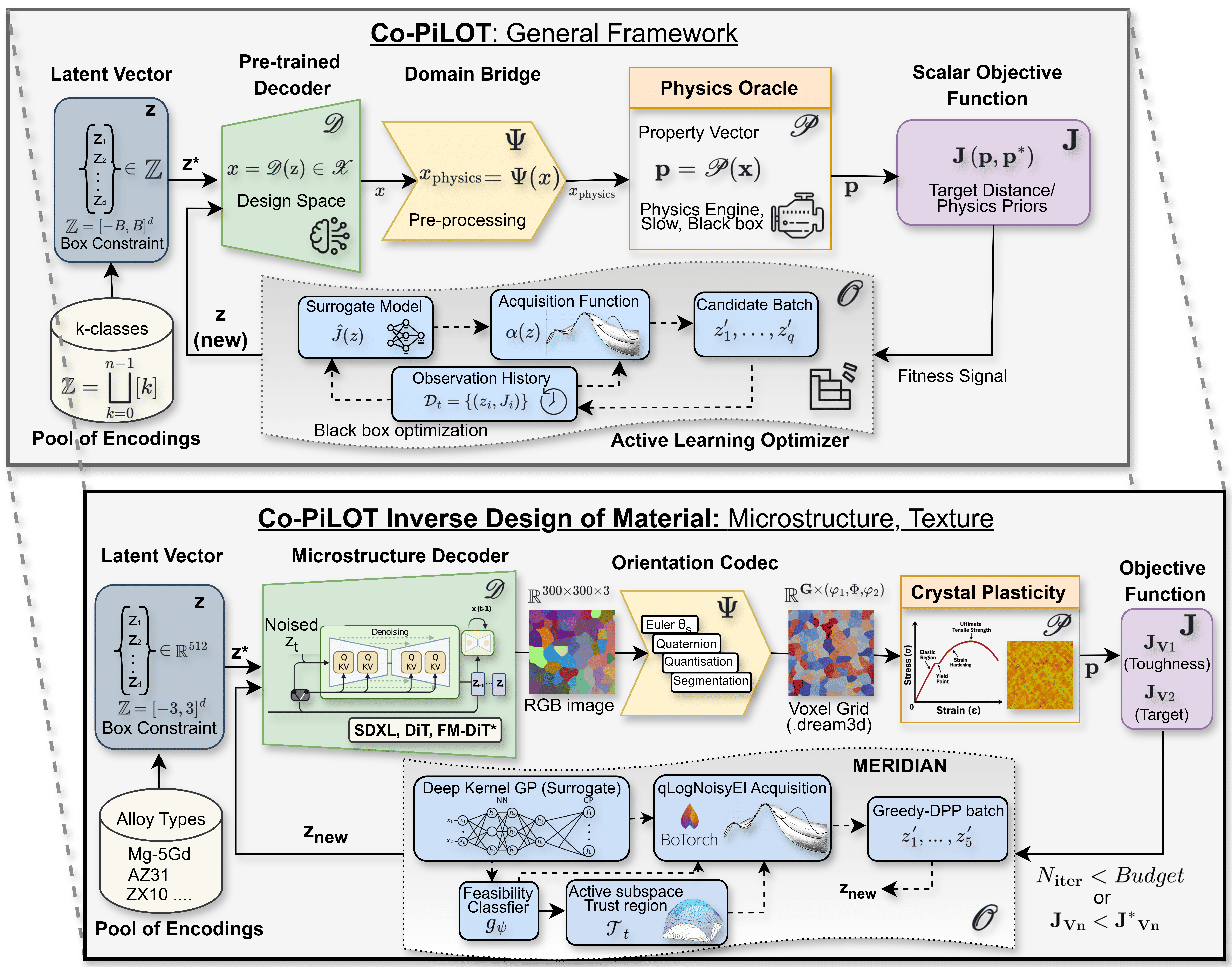}
   \caption{\textbf{\Co{}.} (a)~General framework and its materials instantiation (b): a Mg-alloy latent $z^\star \in \Z$ is passed through a decoder $\Dec$ like \textsc{DiT} or \textsc{FM-DiT}, bridged via an orientation codec ($\Psi$) to a \texttt{.dream3d} grid, scored by a crystal plasticity oracle $f$, and reduced to a fitness $J$ ($J_{V1}/J_{V2}$) that the \meridian{} optimizer $\Opt$ feeds back as $z_\mathbf{new}$ for the next iteration. Physics enters via the simulator-based objective and feasibility model, and via codec-encoded training data (Sec.~\ref{sec:method}).}
  \label{fig:hero}
\end{figure}

\section{Related work}
\label{sec:related}

\textbf{High-dimensional latent optimization.} Latent-space optimization was popularized by \citet{gomez2018}, who showed that a learned continuous representation can make structured design amenable to gradient-based and Bayesian optimization. Follow-up latent-BO methods add weighted retraining, trust regions, and constrained or multi-objective variants~\citep{tripp2020,maus2022,stanton2022,zeng2024antibody}, but they are largely evaluated with cheap neural property predictors rather than expensive physical simulators. High-dimensional BO methods such as \turbo{}, \baxus{}, and \saasbo{} improve sample efficiency in large ambient spaces,\citep{eriksson2019,papenmeier2022,eriksson2021} yet they do not exploit generative priors or explicitly address simulator divergence and structured feasibility, which are central in our setting. Classical black-box optimizers such as CMA-ES remain strong generic baselines~\citep{hansen2001}, although full covariance adaptation scales poorly in high dimensions; we include it in Sec.~\ref{sec:e2}.

\textbf{Generative inverse design of materials.} Generative models have become a standard tool for inverse design in materials and related physical systems~\citep{Lu_2022}. Early materials work used GANs to generate microstructures and couple them to regressors or Bayesian optimization~\citep{yang2018,cang2018}, while more recent studies perform inverse design of dual-phase steel microstructures with generative models and BO~\citep{KusampudiDiehl2023}, inverse design of spinodoid architected materials in the small-data regime via BO~\citep{Rassloff2026}, or use diffusion models for nonlinear mechanical metamaterials and multi-material structures~\citep{ParkKushwaha2024,Zheng2026}. For polycrystals, even representing microstructure for design remains open~\citep{bostanabad2018}, and inverse microstructure and texture design rely on informative low-dimensional priors or conditional normalizing flows in active-learning loops~\citep{generale2024,buzzy2025}. In crystals, CDVAE, DiffCSP, and MatterGen show that modern generative models can capture complex structure distributions~\citep{xie2022,jiao2023,zeni2025}, but these methods are primarily built for sampling or one-shot conditional generation rather than iterative optimization under a strict expensive-oracle budget.

\textbf{Black-box optimization around generators.} Our work is closest in spirit to methods that place an outer optimization loop around a pretrained generator, as in latent BO for molecules~\citep{gomez2018,tripp2020}, photonic inverse design with neural generators and EM solvers~\citep{Molesky2018,Zhou2021,Yeung2023}, and GAN-based topology optimization with FEM~\citep{nie2020,ParkKushwaha2024}. Diffusion black-box optimizers (DDOM~\citep{krishnamoorthy2023ddom}) and reward fine-tuning of diffusion models (DDPO~\citep{black2024ddpo}, SEIKO~\citep{uehara2024seiko}) assume offline data or far larger query budgets; optimizing over learned models can exploit adversarial modes~\citep{wu2024cindm}, and coupling generative models to PDE physics remains frontier work~\citep{bastek2025pidm}. The key gap is that these settings typically rely on cheaper solvers, smoother objectives, or direct conditional generation. \Co{} instead targets the harsher regime of constrained latent optimization with a slow, occasionally divergent physics solver, and therefore emphasizes calibrated surrogates, feasibility modeling, and robust search in pretrained latent spaces.

\section{Method}
\label{sec:method}

The central act of \Co{} is the tight coupling of a pretrained microstructure decoder $\Dec$ (Sec.~\ref{sec:enc-dec}) with a physics-informed optimizer $\Opt$ (Sec.~\ref{sec:pipeline}) via an orientation codec ensuring crystallographic validity (Sec.~\ref{sec:codec}). Operating hand-in-hand inside a single loop (Fig.~\ref{fig:hero}), the decoder turns a latent vector $z{\in}\Z$ into a candidate design, while the optimizer proposes the next $z$ to query based on a scalar fitness $J$ returned by an oracle simulator $f$ (Sec.~\ref{sec:optimizers}).

\textbf{Problem formulation}. Let $\X$ denote a structured design space (microstructures, molecules, \dots), $f : \X \to \R^k$ an expensive oracle returning a property tuple $p = f(x) \in \R^k$, $p^\star \in \R^k$ a target vector, $J : \R^k \to \R$ a scalar objective combining target distance and physics priors, and $g : \X \to \R^m$ feasibility constraints. Given a pretrained encoder--decoder pair $(\Enc, \Dec)$ with $\Dec : \R^d \to \X$, \Co{} maps $\X$ into the latent box $\Z = [-B, B]^d$ inherited from the latent prior and solves
\begin{equation}
  z^\star \;=\; \arg\max_{z \in \Z}\; J\bigl(f(\Dec(z));\, p^\star\bigr)
                  \;-\; \lambda \cdot \mathrm{ReLU}\!\bigl(g(\Dec(z))\bigr),
  \qquad x^\star = \Dec(z^\star).
  \label{eq:latent-inverse-design}
\end{equation}
This reformulation has three structural advantages.
(i)~Deliberate validity as $\Dec(z)$ is always a plausible design.
(ii)~Box constraint gives a natural prior $\Z$.
(iii)~Regularised search space as the decoder restricts the optimizer to a learned manifold, resulting in structured modeling target than direct search over the raw $\X$. The materials instantiation (Fig.~\ref{fig:hero}b) dictates the anatomy of this section.

\textbf{Where physics enters \Co{}.} ``Physics-informed'' follows the physics--ML taxonomies of \citet{vonrueden2021informed,karniadakis2021piml}, not PINN-style residual losses. Physics enters at two levels (Fig.~\ref{fig:hero}b). \emph{(1)~Physics-informed objective:} $J$ is computed from a crystal-plasticity simulation at every iteration ($J_{V2}$ measures distance to the target properties, $J_{V1}$ adds a ductility floor, both penalise unphysical grain counts), and \meridian{}'s feasibility model learns simulator divergence (Secs.~\ref{sec:pipeline},~\ref{sec:meridian}); this is knowledge integration via the learning objective~\citep{vonrueden2021informed}. \emph{(2)~Physics-augmented learned components:} the encoder--decoder is trained on codec-encoded EBSD maps whose every pixel is a valid HCP orientation, so the latent manifold is a manifold of physically realisable microstructures, and \meridian{}'s surrogate is warm-started on $100$ crystal-plasticity simulations rather than a cold random prior (Secs.~\ref{sec:codec},~\ref{sec:e2}).

\subsection{Implementation of Encoder--Decoder ($\Dec$) }
\label{sec:enc-dec}

\Co{} is built around a continually \emph{pre-trained} stack: a ViT-H/14 encoder~\citep{radford2021clip,schuhmann2022laion5b} and diffusion decoders- SDXL~\citep{podell2023sdxl}, SD3.5~\citep{esser2024sd3}, and DiT-XL/2~\citep{peebles2023dit}, that are pre-trained on massive natural-image datasets. We posit this prior is very effective for microstructure reconstruction on a very thin bottlneck $z$. Even a $d{=}512$ bottleneck (${\sim}1500{\times}$ compression) recovers grain morphology, crystallographic colour, and property statistics that are compareable to a from-scratch decoder (Sec.~\ref{sec:e1}).

\begin{figure}[t]
  \centering
  \includegraphics[width=\linewidth]{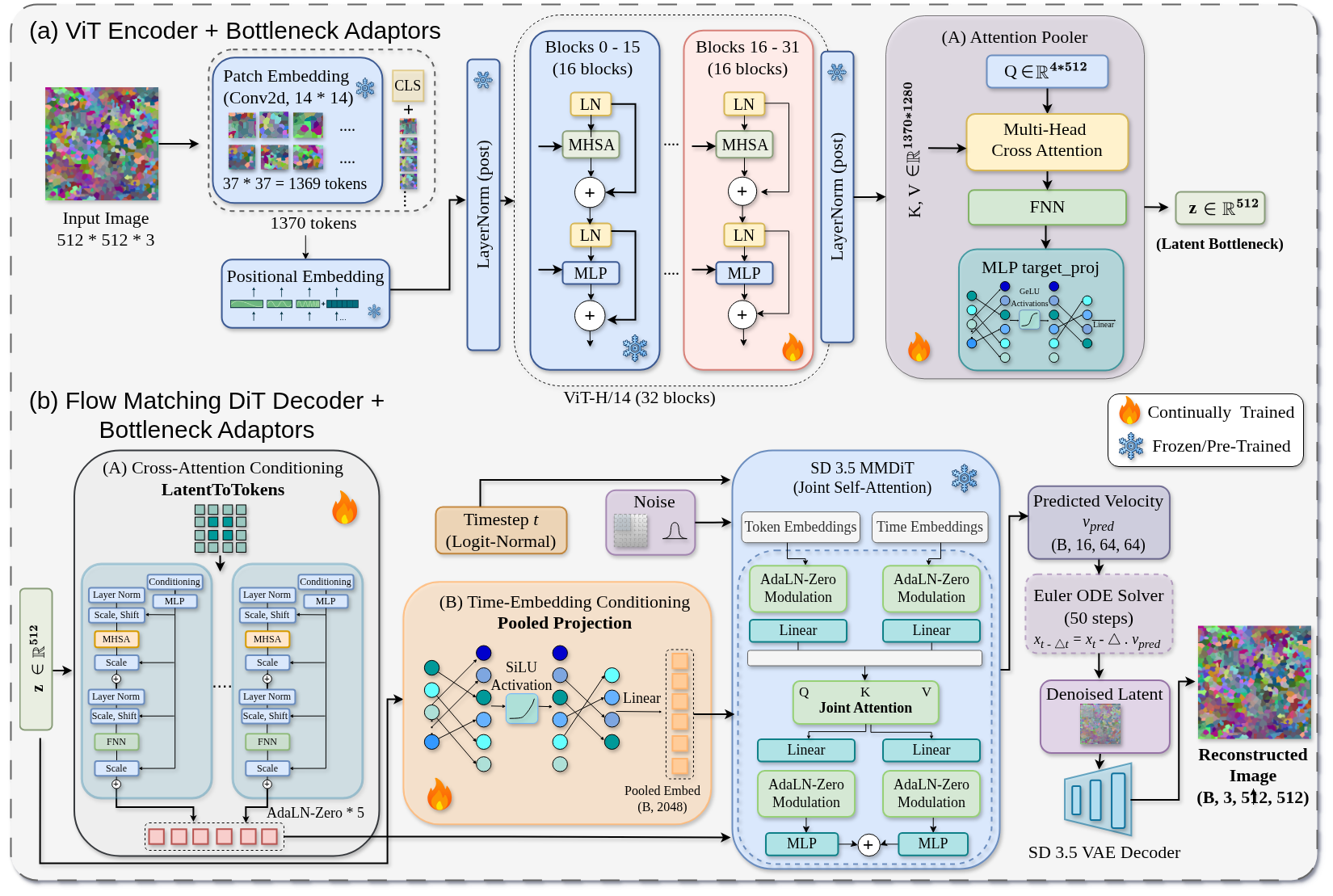}
  \caption{(a)~\textbf{Encoder $\Enc$}, ViT-H/14 trunk $\to$ $\textsc{Attention Pooler}$ with $z\!\in\!\R^{d}$. (b)~\textbf{Decoder $\Dec$}, Frozen SD3.5 MMDiT conditioned via a $\textsc{LatentToTokens}$ adaptor and $\textsc{Pooled Projection}$.}
  \label{fig:archs}
\end{figure}

\textbf{A single shared encoder architecture.} A ViT-H/14 trunk maps $512^2{\times}3$ orientation-as-RGB microstructure images to $1369$ patch tokens plus \texttt{[CLS]} token. 2B weights are reused with the lower $16$ blocks frozen and the upper $16$ unfrozen (Fig.~\ref{fig:archs}(a)), retaining the natural-image prior while letting the top of the trunk specialise to EBSD imagery. The \texttt{[CLS]}-only readout is replaced by a four-query \textsc{Attention Pooler} (cross-attention over all $1370$ tokens, query self-attention, per-query FFN, then merged) followed by a two-layer MLP that maps to bottleneck $z\in\R^{d}$, with $d{=}512$ as the headline; $d\in\{768,1024\}$ are reported as the bottleneck-capacity ablation in Sec.~\ref{sec:e1}.

\textbf{Three diffusion decoders} are each conditioned solely on the bottleneck, $z$. The pre-trained backbones are frozen and lightweight adaptors are introduced that translate $z$ into the native conditioning format of each model (Fig.~\ref{fig:archs}b). \textbf{FM-DiT} (frozen SD3.5 MMDiT~\citep{esser2024sd3}): We introduce a $\textsc{LatentToTokens}$ adaptor using $16$ queries and $5$ QK-normed AdaLN-Zero blocks~\citep{peebles2023dit}. This generates $16{\times}4096$ tokens for the joint attention layers, while a $\textsc{Pooled Projection}$ head provides the $2048$-D global embedding. \textbf{SDXL} (frozen denoising UNet~\citep{podell2023sdxl}): The latent $z$ is mapped to $77{\times}2048$ tokens via $3$ AdaLN-Zero blocks. We initialize the outer layers with $0.10$ Xavier gain, as standard zero-initialization collapses the conditioning heads. A $\textsc{Pooled Projection}$ head maps $z$ to the $1280$-D embedding, and we apply classifier-free guidance~\citep{ho2022cfg}. \textbf{DiT} (DiT-XL/2~\citep{peebles2023dit}): Unlike the others, this model is \emph{fully trainable}, as freezing the backbone fails during the $256{\to}512$ resolution transfer. We repurpose the DiT class-embedding slot to inject $z$ using a $\textsc{ClassProj}$ head. Additionally, $16{\times}1152$ spatial tokens are injected via zero-initialized residuals at blocks $\{0,7,14,21\}$. Architectures for SDXL and DiT are deferred to App.~Fig.~\ref{fig:archs-appendix}. We also include a pixel-space VQGAN~\citep{esser2021vqgan} with a separate codebook as a non-$z$-bottleneck baseline.

\textbf{Joint training}. Encoder, conditioning adaptors, and backbone (for ViT-DiT) are trained jointly under a composite of latent-space generative terms, a latent-alignment regulariser, and a pixel-space frequency term that is gated to small denoising times $t$. The loss function per batch, with $\hat x_0$ as the clean-latent estimate, $z$ as the bottleneck, and $z'$ as its augmentation-twin embedding is given as,
\begin{equation}
  \mathcal{L} \;=\; \lambda_{\mathrm{diff}}\,\mathcal{L}_{\mathrm{diff}}
              + \lambda_{\mathrm{rec}}\,\mathcal{L}_{\mathrm{rec}}
              + \lambda_{\mathrm{con}}\,\mathcal{L}_{\mathrm{con}}
              + \lambda_{\mathrm{vic}}\,\mathcal{L}_{\mathrm{vic}}
              + \lambda_{\mathrm{freq}}\,\mathcal{L}_{\mathrm{freq}},
  \label{eq:composite-loss}
\end{equation}
where $\mathcal{L}_{\mathrm{diff}}$ is the backbone's native objective --- the rectified-flow velocity loss $\lVert v_\theta(x_t,t,z) - (\epsilon - x_0)\rVert_2^2$ for ViT-FMDiT~\citep{lipman2023flowmatching} and the standard MSE loss for ViT-SDXL and ViT-DiT; $\mathcal{L}_{\mathrm{rec}} = \mathbb{E}_t\bigl[(1{-}t)^2\,\lVert\hat x_0 - x_0\rVert_2^2\bigr]$ is a timestep-weighted clean-latent reconstruction term that emphasises low-noise steps; $\mathcal{L}_{\mathrm{con}}$ is an InfoNCE-style contrastive loss over $\{z, z'\}$ pairs; $\mathcal{L}_{\mathrm{vic}}$ is a VICReg variance--covariance regulariser~\citep{bardes2022vicreg} that keeps the bottleneck non-collapsed; $\mathcal{L}_{\mathrm{freq}} = \lVert \log(1{+}|\mathcal{F}(\hat I)|) - \log(1{+}|\mathcal{F}(I)|)\rVert_1$ is an FFT-amplitude pixel loss with a radial high-frequency mask, evaluated on a small subsample of the batch. The five $\lambda$s follow one fixed \texttt{LossWeightScheduler} schedule, reused for every ViT-FMDiT model and never tuned per experiment: $\lambda_{\mathrm{diff}}$ rises $0.5{\to}1.0$ while $\lambda_{\mathrm{rec}}$ and $\lambda_{\mathrm{con}}$ decay, and the two regularisers enter with a delayed onset (App.~\ref{app:enc-dec}); perturbing it moves held-out PSNR by at most $0.09$\,dB (Tab.~\ref{tab:lambda-sensitivity}). A two-phase schedule freezes the encoder for the first seven epochs and unfreezes its last $16$ blocks at epoch $8$, in step units to absorb the topology change when the optimizer is rebuilt and the cosine schedule recomputed. More details on the implementations can be found in App.~\ref{app:enc-dec}.

\subsection{Orientation codec ($\Psi$)}
\label{sec:codec}

\textbf{Why a dedicated codec?} A generic pretrained image decoder emits an RGB tensor; a crystal-plasticity solver expects an HCP orientation field. The codec is the enabling bridge $\Psi$ in Fig.~\ref{fig:hero}(b): an invertible map between HCP orientation fields and 8-bit RGB images that unlocks the decoder--optimizer loop in Sec.~\ref{sec:optimizers} to use any image-domain generator as a microstructure prior. Naively mapping Bunge--Euler orientations $(\varphi_1, \Phi, \varphi_2)$ to RGB triples corrupts the any image-domain decoder data in three ways: periodic wrap-around at $0/2\pi$ becomes ringing; For Hexagonically Closed Pack (HCP) fundamental zone, per-grain folding maps physically adjacent grains to opposite quaternions, becoming high-frequency color edges which blurs the decoder output; and recovering the $q_w$ (quaternions) produces NaNs whenever decoder noise pushes negative square-root. The codec resolves all three through a 5-step encode and a closed vectorised decode (Alg.~\ref{alg:codec}).

\textbf{Five-step pipeline}. Each per-grain orientation is converted to a unit quaternion $q\in S^3$~\citep{zhou2019} under HCP point-group $D_6$ ($|\mathcal{S}_{\mathrm{HCP}}|=12$). The encode then proceeds:
\textbf{(1)~Continuous unfolding:} breadth-first-search (BFS) over the grain-adjacency graph replaces each grain's quaternion by the symmetry equivalent closest to its already-visited parent (Eq.~\ref{eq:bfs-step}), so RGB distance between adjacent grains is proportional to their true misorientation, no artificial fundamental-zone jumps.
\textbf{(2)~Class-level anchor:} a single per-class anchor $\bar q_c$ is precomputed via the eigenvalue method of~\citet{markley2007} from $M{=}50$ file-level means; using a class-level anchor enables decoder generated images without per-sample metadata.
\textbf{(3)~Frame centring.} $q_g\leftarrow \bar q_c^{-1}\cdot q_g$ concentrates the distribution near identity, away from the $q_w{=}0$ equator where the double-cover sign flip would reintroduce discontinuities.
\textbf{(4)~Stereographic projection.} $S = q_{xyz}/(1+q_w) \in [-1,1]^3$ has the closed-form rational inverse in Eq.~\ref{eq:istereo} (no square roots, numerically stable on all of $\R^3$).
\textbf{(5)~Quantisation:} $S$ is linearly mapped to $[0, 2^b{-}1]^3$. The per-channel step $\Delta = 2/(2^b{-}1)$ bounds the worst-case angular error at ${\approx}0.78^\circ$. The 8-bit PNG variant is used inside \Co{}, it plugs directly into a pretrained $\Dec$, and its measured ${\sim}0.6^\circ$ mean error (App.~\ref{app:codec-extended}, Tab.~\ref{tab:codec-accuracy}) is far below the grain-boundary threshold~\citep{readshockley1950}. \textbf{Self-segmenting decode} is done at inference where no ground-truth label map is available for a generated image. The decoder recovers grains by linking $4$-connected pixels whose max-channel intensity differs by at most $\tau$ ($\tau{=}1$ for $8$-bit, $\tau{=}50$ for $16$-bit image) via a single sparse-graph connected-components pass~\citep{groeber2014}.

\begin{algorithm}[t]
\caption{Orientation codec --- encode (Dream3D $\to$ image) and decode (image $\to$ Dream3D)}
\label{alg:codec}
\begin{algorithmic}[1]
\Statex \textbf{Encode}\,$(\mathcal{G}, \{q_g\}, \mathcal{N}{=}(\mathcal{V},\mathcal{E}), \bar{q}_c)$
\State $g^\star \gets \arg\max_{g \in \mathcal{V}} |\{(i,j):\mathcal{G}[i,j]{=}g\}|$; $q_{g^\star} \gets \arg\max_{s,\sigma}\, \sigma\langle s\cdot q_{g^\star},\bar{q}_c\rangle$
  \Comment{root + anchor}
\For{$(g_p, g_c) \in \mathcal{E}$ from $g^\star$}
  \State $q_{g_c} \gets \arg\max_{s,\sigma}\, \sigma\langle s\cdot q_{g_c},\, q_{g_p}\rangle$
         \Comment{Eq.~\eqref{eq:bfs-step}}
\EndFor
\State $q_g \gets \bar{q}_c^{-1}\cdot q_g$ with $q_{g,w}\!\ge\!0$; $S_g \gets q_{g,xyz}/(1+q_{g,w})$
       \Comment{Eq.~\eqref{eq:stereo}}
\State \Return $\mathrm{img}[i,j] = \bigl\lfloor (S_{\mathcal{G}[i,j]}+1)/2\,\cdot\,(2^b{-}1) \bigr\rceil$
\Statex \textbf{Decode}\,$(\mathrm{img}, \bar{q}_c)$
\State $S \gets 2\,\mathrm{img}/(2^b{-}1) - 1$; $q_w \gets (1-\norm{S}^2)/(1+\norm{S}^2)$, $q_{xyz} \gets 2S/(1+\norm{S}^2)$
       \Comment{Eq.~\eqref{eq:istereo}}
\State $q \gets \bar{q}_c \cdot q$, then fold: $q \gets \arg\max_{s}\,[s\cdot q]_w$
       \Comment{HCP fundamental zone}
\State $\mathcal{G} \gets \mathrm{conn\_comp}_{4}(\mathrm{img};\,\tau)$
       \Comment{$\tau{=}1$ for $b{=}8$}
\State \Return DAMASK-ready \texttt{.dream3d} with per-pixel Euler
\end{algorithmic}
\end{algorithm}

\subsection{Crystal plasticity simulation pipeline and objectives ($f\circ\Dec$, $J$)}
\label{sec:pipeline}

With the codec in place, the materials oracle composes into
\begin{equation*}
  z \xrightarrow{\;\Dec\;} \text{PNG} \xrightarrow{\;\text{codec}\;} \text{.dream3d} \xrightarrow{\;\damask\;} \text{HDF5} \xrightarrow{\;\text{Hollomon}\;} p \xrightarrow{\;J\;} \R,
\end{equation*}
where the mechanical property tuple $p = (\sigy,\,n,\,K,\,\sigu,\,\varepsilon_u)$ collects the $0.2\%$-offset yield stress $\sigy$, the Hollomon hardening exponent $n$, strength coefficient $K$, the ultimate tensile stress $\sigu$, and the uniform strain $\varepsilon_u$ at $\sigu$. The grain table emitted by the codec (Sec.~\ref{sec:codec}) — per-grain median Euler angles, voxel counts, phase IDs — is written as a \dreamthreed{} representative volume element (RVE) polycrystal that \damask{} consumes directly. Each candidate is driven through the \damask{} \texttt{Grid} solver, an FFT-based spectral scheme for periodic RVEs~\citep{roters2019}, under uniaxial tension along the extrusion direction at a quasi-static strain rate ($\dot\varepsilon = 10^{-3}\,\mathrm{s}^{-1}$, $25\%$ total nominal strain). The constitutive law is the HCP phenomenological power law of \citet{roters2019} with the Mg-alloy AZ31-calibrated parameter set covering tensile twinning; basal, prismatic, and pyramidal slip deformation mechanisms explained in App.~\ref{app:damask}). Non-converged runs — a small minority, primarily on degenerate decoded grains — are excluded from the surrogate's training set in all optimizers.

\textbf{Objective:} $J(p)$ has two complementary forms: a \emph{toughness-weighted} strength score $J_{V1}$ that simultaneously rewards high yield $\sigy$ and high plastic work to UTS (proxied by $\sigu\,\varepsilon_u$), and a \emph{target-driven} score $J_{V2}$ that penalises weighted distance to a user-supplied target $p^\star=(\sigy^\star, n^\star, K^\star, \sigu^\star)$,
\begin{equation}
  J_{V1}(p) = \Bigl(\tfrac{\sigy}{\sigy^{\mathrm{ref}}}\Bigr)^{\!\alpha}\!\Bigl(\tfrac{\sigu\,\varepsilon_u}{T^{\mathrm{ref}}}\Bigr)^{\!\beta}
             - \lambda_n\,\mathrm{ReLU}(n_{\min}{-}n),
  \qquad
  J_{V2}(p; p^\star) = -\sqrt{\textstyle\sum_{p\in\mathcal{I}} w_p\bigl(\tfrac{p - p^\star}{|p^\star|}\bigr)^{\!2}}.
  \label{eq:objectives}
\end{equation}
The $n_{\min}$ floor in $J_{V1}$ rejects brittle solutions; both add a saturating grain-count band penalty (App.~\ref{app:damask}) since the constitutive law is grain-size insensitive and the optimizer would otherwise reach the target via a few large favourably-oriented grains or via speckle decodes with thousands of grains. The constants in Eq.~\eqref{eq:objectives} ($\alpha$, $\beta$, $w_p$, $p^\star$) are not tuned: they encode the design goal, so changing them changes the question being asked rather than how well it is answered.

\subsection{Optimizer: \meridian{} (Ours) ($\Opt$)}
\label{sec:optimizers}
\label{sec:meridian}

We deploy \meridian{} (\emph{Manifold-Embedded Robust Inverse Design via Iterative Acquisition Networks}) as the primary optimizer and benchmark it against seven baselines under the same box, seed cache, and budget: \dante{}~\citep{wei2025} (MLP surrogate with tree exploration), \turbo{}~\citep{eriksson2019} (GP with a single trust region), \baxus{}~\citep{papenmeier2022} (GP in a random subspace), CMA-ES~\citep{hansen2001} and DDOM~\citep{krishnamoorthy2023ddom} as optimizer-level comparisons, and SEIKO~\citep{uehara2024seiko} and DDPO~\citep{black2024ddpo}, which fine-tune the decoder itself, as framework-level comparisons. Implementations and hyperparameters (Tabs.~\ref{tab:opt-comparison},~\ref{tab:hparams-opt}) are in App.~\ref{app:baselines}.

\textbf{Why a dedicated optimizer?} Three properties specific to \Co{} shapes the design. The oracle is expensive ($\sim 15$\,min/sim) and a non-trivial fraction of evaluations fail at decode, codec, or solver gates, so the optimizer must model both where the response is uncertain and where it returns a finite value. Pre-trained image-domain decoders place their training mass on a thin spherical shell of the latent box (e.g.\ $\| z\| {=} 22.07{\pm}0.05$ for ViT-FMDiT-$512$), so any perturbations in $\R^{512}$ leave this shell and waste simulator calls. \dante{}'s point-estimate MLP and visit-count tree miss all three; the GP-trust-region baselines satisfy uncertainty but neither manifold geometry nor batch diversity. \meridian{} retains the iterative ``surrogate $\to$ candidate cloud $\to$ diverse batch'' skeleton common to other three optimizers and replaces every component on which it breaks for this problem (Tab.~\ref{tab:opt-comparison}).

\begin{algorithm}[t]
\caption{\meridian{}, one outer round per iteration $t = 1, \dots, T$.}
\label{alg:meridian}
\begin{algorithmic}[1]
\Require{Box $\Z$, batch size $q$, oracle $f$, decoder $\Dec$, class pool $\mathcal P$, seed cache ${\Dset}_{0}$.}
\State ${\Dset} \gets {\Dset}_{0}$;\quad $L \gets L_{\mathrm{init}}$;\quad $r,\,\mathrm{plat} \gets 0$
\For{$t = 1, \dots, T$}
  \State fit trunk, feasibility head, and GP on $\Dset$ \hfill \emph{(surrogate)}
  \State $w \gets \mathrm{AS}(\nabla \hat\mu)$ if $t \ge 3$, else $\mathrm{PCA}(\mathcal P)$ \hfill \emph{(subspace)}
  \State $z_c \gets \mathrm{centroid}_{\mathrm{top}K}(\Dset)$;\quad
    $Z_{\mathrm{cand}} \gets \Pi_{\mathrm{shell}}$,\; $U \sim \mathrm{Sobol}$ \hfill \emph{(cloud)}
  \State $\alpha(z) \gets \mathrm{qLogNEI}(z;\mu,\sigma)\, g_\psi(z)$ for $V2$ \hfill \emph{(acquisition)}
  \State $S \gets \mathrm{top}_{256}(Z_{\mathrm{cand}}, \alpha)$;\quad
    $Z_{\mathrm{batch}} \gets \mathrm{GreedyDPP}(S, q;\, k_{(\phi,z)})$ \hfill \emph{(batch)}
  \State $\Dset \gets \Dset \cup \{(z, f(\Dec(z)))\}_{z \in Z_{\mathrm{batch}}}$;\quad update $L$ and $\mathrm{plat}$
  \State \textbf{if} $L < L_{\min}$ or $\mathrm{plat} \ge K_{\mathrm{plat}}$ \textbf{then} restart from $\mathcal P$ or mini-MCTS
\EndFor
\State \Return $z^\star = \arg\max_{c_i=1} y_i$ and $\Dec(z^\star)$.
\end{algorithmic}
\end{algorithm}

\textbf{One round of \meridian{}.} Each outer iteration begins by refitting the surrogate: a shared MLP feature map $\phi_\theta : \R^{512} \to \R^{16}$ is trained jointly with a logistic feasibility classifier $g_\psi$ (on \emph{all} observations, so the failure signal is preserved), after which an exact ARD-Mat\'ern-$5/2$ GP is fit on the feasible subset on top of $\phi_\theta$ to supply calibrated $(\mu, \sigma^2)$. From the surrogate we read off an importance-weighted axis vector $w \in \R^{512}$ via the active subspace of $\hat C = \E[\nabla\hat\mu\,\nabla\hat\mu^\top]$ (PCA of the class-conditional pool $\mathcal P$ during cold start)~\citep{constantine2015}. We then anchor the round on the shell-projected centroid of the top-$K$ feasible incumbents and draw a Sobol cloud around it whose perturbations are scaled by the trust-region edge $L_t$ and the importance weights $w$ on a sparse axis mask, so that proposals respect the manifold's anisotropy without ignoring rare-direction pockets, and then project the cloud onto the adaptive shell $\| z\| \in [\mu_{\| X\|} \pm k_\sigma\, \sigma_{\| X\|}]$, which is measured from the feasible data each round. Acquisition combines a feasibility-gated qLogNoisyExpectedImprovement, $\alpha_{\mathrm{GP}}(z) = \mathrm{qLogNEI}(z; \mu, \sigma)\, g_\psi(z)$, a Monte-Carlo EI term, for the target-driven $V2$ objective, computed through heteroscedastic property heads on $\phi_\theta$. The top-$256$ candidates ranked by $\alpha$ are passed to a greedy quality-weighted DPP~\citep{kulesza2012} with a hybrid $(\phi, z)$ kernel that returns a diverse batch of proposals; the batch is then evaluated through $\Dec \to \damask{}$, $L_t$ updated under success/failure cadence, and a restart fired whenever $L_t$ collapses or best plateau runs overflow $K_{\mathrm{plat}}$ rounds. \meridian{}, in this configuration with hyperparameters and more explanation are in App.~\ref{app:meridian-details}.

\section{Experiment 1: Microstructure reconstruction (E1)}
\label{sec:experiments}
\label{sec:e1}

The encoder--decoder models described in Sec.~\ref{sec:enc-dec} are trained once and then frozen for all downstream optimization experiments in Sec.~\ref{sec:e2}. Expanding on the data corpus of \citet{guru2025}, we assemble $81{,}756$ training and $9{,}084$ held-out test RGB microstructures; this expansion was achieved through geometry and physics-aware oversampling, the details of which are outside the scope of this paper and will be documented in a future publication. All the models from Tab.~\ref{tab:reconstruction} are trained on $4\times$H200 GPUs, and a standard $30$-epoch run on the training split takes roughly $12$ hours. The VQGAN reference, FM-DiT at $z\in\{512,768,1024\}$, and SDXL all use epoch-$30$ checkpoints; DiT is substantially heavier because its denoiser backbone is unfrozen, carrying about $1.17$B trainable parameters versus roughly $381$M--$391$M for the FMDiT family, and the comparison therefore uses the latest available epoch-$20$ checkpoint from the same training regime. We score reconstruction quality with three computer-vision metrics and two material science metrics. FID is the distribution-level realism score, MS-SSIM captures multi-scale structural similarity, and LPIPS captures perceptual distance~\citep{heusel2017gans,Wang2003MultiscaleSS,zhang2018perceptual}. For materials fidelity, we report $G_{\mathrm{m}}$ for grain recovery (fraction of original grains recovered at IoU $\ge 0.3$, capturing grain size and aspect-ratio agreement) and $\Delta_{\mathrm{ori}}$ for crystallography (mean disorientation between IoU-matched per grains per-class mean quaternions).

\begin{figure*}[htbp]
  \centering
  \includegraphics[width=\linewidth]{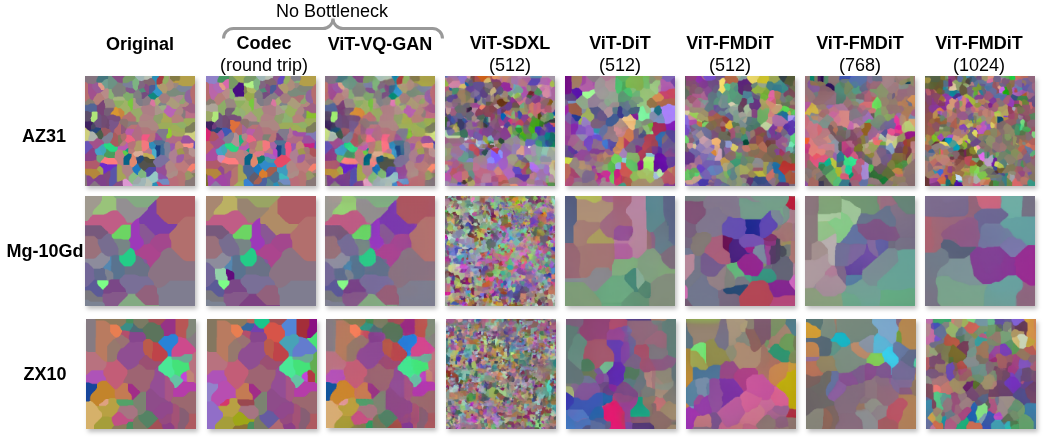}
  \caption{\textbf{Qualitative Reconstructions} on AZ31 and Mg-10Gd. More examples in Fig.~\ref{fig:reconstruction-extended}}
  \label{fig:reconstruction}
\end{figure*}

\begin{table*}[htbp]
  \caption{\textbf{Reconstruction quality and microstructural fidelity on the test set} ($n = 9{,}084$).}
  \label{tab:reconstruction}
  \centering
  \footnotesize
  \setlength{\tabcolsep}{4pt}
  \resizebox{0.95\textwidth}{!}{%
  \begin{tabular}{lc ccccc}
    \toprule
    \textbf{Decoder} & \textbf{$z$} & \textbf{FID $\downarrow$} & \textbf{MS-SSIM $\uparrow$} & \textbf{LPIPS $\downarrow$} & \textbf{$G_{\mathrm{m}}$ $\uparrow$} & \textbf{$\Delta_{\mathrm{ori}}$ $\downarrow$} \\
    \midrule
    \multicolumn{7}{l}{\textit{No bottleneck}} \\
    \textbf{VQGAN} & -- & 35.63 & $0.876 \pm 0.12$ & $0.132 \pm 0.10$ & $0.747 \pm 0.16$ & $27.619 \pm 10.01$ \\
    \textbf{Codec} & -- & 0.95 & $0.449 \pm 0.28$ & $0.368 \pm 0.15$ & $0.926 \pm 0.08$ & $7.200 \pm 4.26$ \\
    \midrule
    \textbf{SDXL} & 512 & 108.28 & $0.077 \pm 0.05$ & $0.650 \pm 0.08$ & $0.370 \pm 0.15$ & $56.467 \pm 8.41$ \\
    \textbf{DiT} & 512 & $\mathbf{23.19}^{*}$ & $0.147 \pm 0.18$ & $0.603 \pm 0.08$ &  $0.486 \pm 0.12$ & $55.363 \pm 10.35$ \\
    \textbf{FM-DiT} & 512 & 27.86 & $0.169 \pm 0.20$ & $0.578 \pm 0.07$ &$\mathbf{0.501 \pm 0.11}^{*}$ & $\mathbf{53.921 \pm 11.68}^{*}$ \\
    \textbf{FM-DiT} & 768 & 27.86 & $\mathbf{0.178 \pm 0.21}^{*}$ & $\mathbf{0.572 \pm 0.07}^{*}$ & $0.474 \pm 0.12$ & $54.318 \pm 11.16$ \\
    \textbf{FM-DiT} & 1024 & 27.08 & $0.174 \pm 0.21$ & $0.574 \pm 0.07$ & $0.475 \pm 0.12$ & $54.353 \pm 10.81$ \\
    \bottomrule
  \end{tabular}%
  }
  \vspace{1pt}
  \parbox{0.9\linewidth}{\footnotesize All bottlenecked decoders share the ViT based encoder; names omit the ViT prefix. $^*$Best bottlenecked decoders.}
\end{table*}

ViT-DiT achieves the best FID, while the ViT-FMDiT family (specifically at $d{=}768$) achieves the best $G_{\mathrm{m}}$—indicating the strongest grain-level recovery in size and shape—alongside the best MS-SSIM, LPIPS, and $\Delta_{\mathrm{ori}}$ among bottlenecked decoders (Tab.~\ref{tab:reconstruction}). ViT-FMDiT's performance is stable across bottleneck size, allowing the optimization loop (Sec.~\ref{sec:e2}) to use a tighter bottleneck without sacrificing fidelity. Conversely, ViT-SDXL struggles to adapt its text-conditioned UNet to a 512-D continuous bottleneck, yielding the worst metrics (FID $\sim\!108$) and visibly washed-out grains in Fig.~\ref{fig:reconstruction}. An unbottlenecked ViT-VQGAN provides an upper bound for reconstruction, while the orientation codec establishes the error floor. The codec's $\Delta_{\mathrm{ori}}$ error is safely within CP tolerances. While the bottlenecked decoders preserve macroscopic morphology and class statistics (e.g., AZ31, Mg-10Gd), their orientation space is non-isometric and symmetry-blind. This introduces a ${\sim}53^\circ$ per-pixel disorientation into the end-to-end round-trip. This residual error acts as a class-preserving reshuffle rather than a physical violation as the decoded grains remain in the HCP fundamental zone, and their texture statistics (quaternion mean, pole-figure spread) matches the target class. Therefore, we carry ViT{-}DiT and ViT{-}FMDiT with $d\in\{512,768,1024\}$ into the \Co{} loop as robust generators of statistically valid microstructures rather than exact reconstructors.

\section{Experiment 2: Materials inverse design (E2)}
\label{sec:e2}
\begin{figure}[t]
  \centering
  \includegraphics[width=\linewidth]{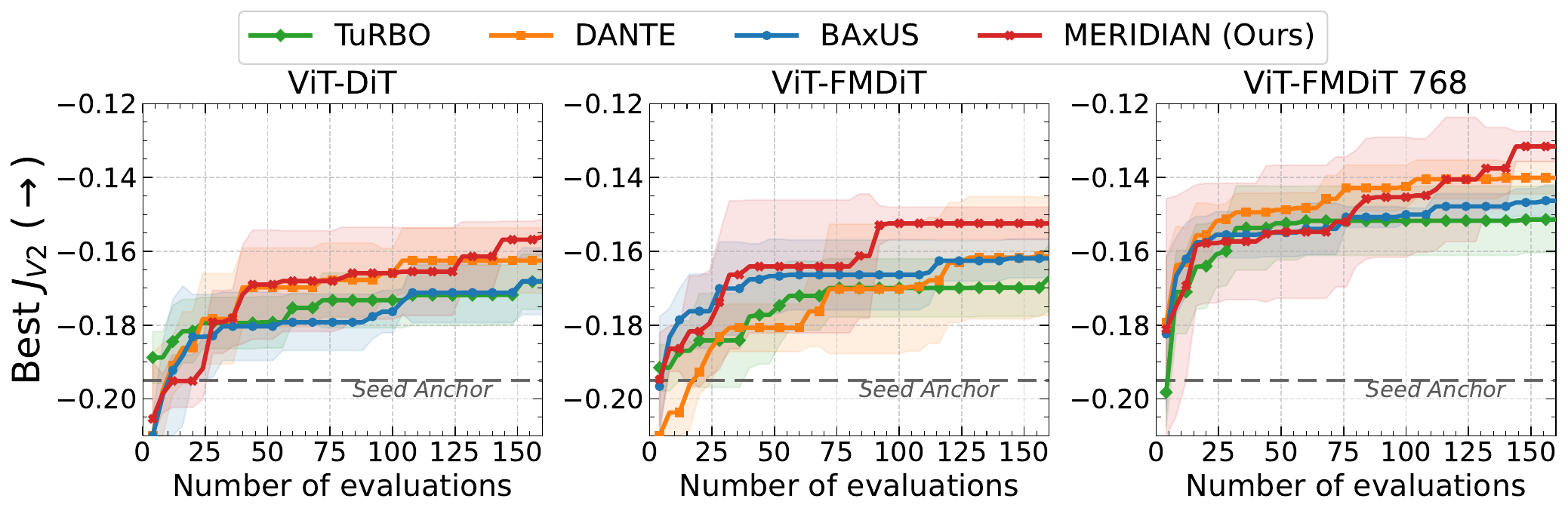}
 \caption{\textbf{Best Mean Target-Driven $J_{V2}$} versus num. of evaluations (batch size 4, $n_{\mathrm{seed}}= 5$).}
  \label{fig:headline}
\end{figure}

\begin{figure}[!t]
  \sbox{\tabvtwobox}{%
    \scriptsize
    \setlength{\tabcolsep}{2.5pt}%
    \renewcommand{\arraystretch}{1.0}%
    \begin{tabular}{@{}llccccc@{}}
      \toprule
      Decoder & Optimizer & $J_{V2}\!\uparrow$ & $\sigy$ & $n$ & $\sigu$ & \# sims \\
      \midrule
      \rowcolor{gray!10} \cellcolor{white}\multirow{4}{*}{ViT{-}DiT}
        & \textbf{\meridian{}}   & $-0.156 \pm 0.005^*$ & $155.9$ & $0.305$ & $313.9$ & $160$ \\
        & \dante{}               & $-0.163 \pm 0.009$ & $155.8$ & $0.305$ & $312.9$ & $104$ \\
        & \turbo{}               & $-0.168 \pm 0.007$ & $154.1$ & $0.301$ & $307.7$ & $152$ \\
        & \baxus{}               & $-0.168 \pm 0.009$ & $154.8$ & $0.303$ & $310.6$ & $152$ \\
      \midrule
      \rowcolor{gray!10} \cellcolor{white}\multirow{4}{*}{ViT{-}FMDiT}
        & \textbf{\meridian{}}   & $-0.152 \pm 0.004^*$ & $156.1$ & $0.302$ & $312.1$ & $96$ \\
        & \dante{}               & $-0.161 \pm 0.016$ & $157.1$ & $0.303$ & $314.4$ & $156$ \\
        & \turbo{}               & $-0.167 \pm 0.009$ & $154.4$ & $0.301$ & $308.4$ & $160$ \\
        & \baxus{}               & $-0.162 \pm 0.005$ & $155.1$ & $0.303$ & $311.1$ & $144$ \\
      \midrule
      \rowcolor{blue!8} \cellcolor{white}\multirow{8}{*}{\shortstack{ViT{-}FMDiT\\[1pt]{\footnotesize 768}}}
        & \textbf{\meridian{}}$^\dagger$ & $\mathbf{-0.132 \pm 0.004}^*$ & $159.1$ & $0.299$ & $314.0$ & $144$ \\
        & \dante{}               & $-0.140 \pm 0.004$ & $157.7$ & $0.301$ & $313.1$ & $140$ \\
        & \turbo{}               & $-0.151 \pm 0.009$ & $158.5$ & $0.306$ & $318.5$ & $148$ \\
        & \baxus{}               & $-0.146 \pm 0.004$ & $157.7$ & $0.305$ & $316.5$ & $156$ \\
        \cmidrule(l){2-7}
        & CMA-ES                 & $-0.144 \pm 0.002$ & $156.6$ & $0.302$ & $313.1$ & $147$ \\
        & DDOM                   & $-0.136 \pm 0.008$ & $157.9$ & $0.301$ & $313.1$ & $123$ \\
        & SEIKO                  & $-0.165 \pm 0.009$ & $153.5$ & $0.304$ & $309.5$ & $133$ \\
        & DDPO                   & $-0.170 \pm 0.003$ & $152.9$ & $0.305$ & $309.7$ & $144$ \\
      \midrule
      \rowcolor{gray!10} \cellcolor{white}\multirow{8}{*}{\shortstack{ViT{-}FMDiT\\[1pt]{\footnotesize 1024}}}
        & \textbf{\meridian{}}   & $-0.150 \pm 0.005^*$ & $157.0$ & $0.304$ & $314.0$ & $152$ \\
        & \dante{}               & $-0.168 \pm 0.004$ & $153.6$ & $0.304$ & $308.8$ & $124$ \\
        & \turbo{}               & $-0.156 \pm 0.009$ & $157.1$ & $0.305$ & $315.0$ & $140$ \\
        & \baxus{}               & $-0.164 \pm 0.008$ & $155.3$ & $0.304$ & $311.6$ & $124$ \\
        \cmidrule(l){2-7}
        & CMA-ES                 & $-0.167 \pm 0.005$ & $154.0$ & $0.307$ & $312.6$ & $144$ \\
        & DDOM                   & $-0.154 \pm 0.009$ & $155.4$ & $0.305$ & $313.0$ & $111$ \\
        & SEIKO                  & $-0.174 \pm 0.011$ & $152.6$ & $0.307$ & $310.4$ & $111$ \\
        & DDPO                   & $-0.176 \pm 0.004$ & $152.7$ & $0.309$ & $312.2$ & $155$ \\
      \bottomrule
    \end{tabular}}%
  \setlength{\figvtwowidth}{0.26\textwidth}%
  \begin{minipage}[c]{\figvtwowidth}
    \centering
    \includegraphics[width=\linewidth]{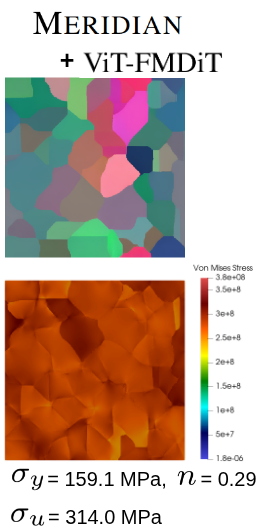}
    \captionof{figure}{Best $J_{V2}$ \\ microstructure and stress field.}
    \label{fig:placeholder-v2}
  \end{minipage}%
  \hfill
  \begin{minipage}[c]{\dimexpr\textwidth-\figvtwowidth-1em\relax}
    \centering
    \captionof{table}{\textbf{Optimization Results:} Target-Driven, $J_{V2}$ ($n_{\mathrm{seed}}= 5$).}
    \label{tab:opt-v2}
    \resizebox{\linewidth}{!}{\usebox{\tabvtwobox}}\\[1pt]
    {\tiny \colorbox{gray!10}{Grey} = ours. \; $^*$ = best per block. \; \colorbox{blue!8}{$^\dagger$} = best overall. \; \# sims = evals to best. \; Below line: added baselines.}
  \end{minipage}
\end{figure}

E2 tests the central claim : Can \Co{} find statistically valid AZ31 microstructures whose simulated stress--strain summaries match a user-specified mechanical target? We use the optimizer suite from Sec.~\ref{sec:optimizers} and the trained decoders from Sec.~\ref{sec:e1}. The target, $p^\star=(\sigy^\star,n^\star,K^\star,\sigu^\star)=(160\,\mathrm{MPa},0.302,604\,\mathrm{MPa},316\,\mathrm{MPa})$ is set. Each cell uses $5$ independent runs and the same $100$ ($z_\mathbf{init}$ <-> \damask{} evaluation) seed cache for warm starting the surrogate model. The optimizer evaluates $40$ iterations with batch size $4$, i.e. $160$ new simulator evaluations per run. Average runtime per evaluation on a $700$ W H200 is $8.46$ hours with a GPU energy estimate of approx $5.93$ kWh across the full sweep. Table~\ref{tab:opt-v2} identifies ViT{-}FMDiT-$768$ with \meridian{} as the strongest target-driven configuration. After 144 evaluations (Fig.~\ref{fig:headline}), it yields the closest overall target match ($J_{V2}=-0.132\pm0.004$, $\sigy=159.1$ MPa, $n=0.299$, $\sigu=314.0$ MPa), successfully falling within the AZ31 material class. This success highlights a key principle: the optimizer and decoder must align. The $768$-D bottleneck offers the ideal trade-off—more expressive than $512$-D, yet dense enough for local surrogate modeling. Conversely, the $1024$-D bottleneck adds expressivity but becomes too sparse under a $160$-evaluation budget, so larger latents do not automatically improve inverse design. This reinforces our core conclusion: successful inverse design requires an optimizer whose inductive biases match the encoder-decoder's latent space (see App.~\ref{app:extended-results} for $J_{V1}$).

\textbf{Extended baselines.} On ViT-FMDiT-$768$/$1024$ we also ran CMA-ES, DDOM, SEIKO, and DDPO under the same protocol (Tab.~\ref{tab:opt-v2}, lower rows). \meridian{} stays best ($-0.132$/$-0.150$); DDOM ($-0.136$/$-0.154$) is within one standard deviation but never ahead, as it models neither feasibility nor trust-region geometry, and surrogate-free CMA-ES trails ($-0.144$/$-0.167$) because it must spend simulator calls to learn the landscape. SEIKO and DDPO, which fine-tune the decoder toward average reward, do not beat the warm-start incumbent within $160$ evaluations, $160$--$360\times$ fewer queries than DDPO's reference configurations. The middle of the ranking reorders between decoders while the top two do not, so the result reflects the method rather than one bottleneck width. Across four decoders and two objectives, \meridian{} is best in all eight studies; batch-rule and sensitivity ablations are in App.~\ref{app:ablations}.

\section{Conclusion}
\label{sec:conclusion}

\textbf{Limitations}. \Co{} is limited by its decoder as it can only search microstructures that the learned prior can express. The decoder is also a class-conditional generator of statistically valid AZ31 microstructures, not a faithful per-pixel orientation reconstructor; although the codec is sub-degree accurate, the encoder--decoder is still non-isometric in crystallographic rotation distance. The \damask{} oracle is HCP phenopower-law plasticity, no damage, isothermal loading, and the loading paths studied here, therefore, inverse-designed microstructures require multi-loading and experimental validation. This paper does \emph{not} show transfer to other polycrystalline materials, property sets, or physical domains, and compares optimizers at a single $160$-evaluation budget; harder inverse-design problems are future work. We are addressing these issues by expanding validation and training $\Enc$--$\Dec$ with orientation-specific losses such as HCP-symmetry-aware grain-level quaternion losses and texture-statistics penalties.

\textbf{Highlights and broader impact}. \Co{} turns expensive simulator-driven inverse design into active search over a learned generative design space; to our knowledge, no similar end-to-end pipeline previously existed for this regime. The decoders are evaluated by FID, MS-SSIM, LPIPS, grain recovery, and disorientation against ground truth (Tab.~\ref{tab:reconstruction}) and \meridian{} against seven optimizers under a matched budget plus batch-rule and control-parameter ablations (Tab.~\ref{tab:opt-v2}, App.~\ref{app:ablations}). The codec makes image-domain generators usable inside a crystal-plasticity loop, and \meridian{} combines calibrated uncertainty, feasibility modeling, manifold-aware proposals, and diverse batches to reduce wasted simulator calls. On AZ31 design, ViT-FMDiT-$768$ with \meridian{} gives the strongest target-driven result and remains effective on the toughness objective. The abstraction may extend to other physical systems with learned priors and slow oracles, which remains to be shown. Code: \urlEncDec{} (encoder--decoder), \urlCodec{} (orientation codec), and \urlMeridian{} (\meridian{}, \damask{} harness).

\begin{ack}
We thank the anonymous reviewers and the area chair for comments that improved the evaluation of this work.

\textbf{Funding.} We acknowledge Helmholtz-Zentrum Hereon for providing computational infrastructure and institutional support. Additional support was provided by the Joachim Herz Stiftung through an Add-on Fellowship.

\textbf{Competing interests.} The authors declare no competing interests.
\end{ack}

{\small
\bibliographystyle{unsrtnat}
\bibliography{references}

@article{cang2018,
title = {Improving direct physical properties prediction of heterogeneous materials from imaging data via convolutional neural network and a morphology-aware generative model},
journal = {Computational Materials Science},
volume = {150},
pages = {212-221},
year = {2018},
issn = {0927-0256},
doi = {https://doi.org/10.1016/j.commatsci.2018.03.074},
url = {https://www.sciencedirect.com/science/article/pii/S0927025618302337},
author = {Ruijin Cang and Hechao Li and Hope Yao and Yang Jiao and Yi Ren}
}

@book{constantine2015,
  author    = {Paul G. Constantine},
  title     = {Active Subspaces: Emerging Ideas for Dimension Reduction in Parameter Studies},
  year      = {2015},
  publisher = {Society for Industrial and Applied Mathematics},
  series    = {SIAM Spotlights},
  volume    = {2},
  address   = {Philadelphia},
  isbn      = {978-1-61197-385-3},
  doi       = {10.1137/1.9781611973860},
  url       = {https://doi.org/10.1137/1.9781611973860}
}

@inproceedings{eriksson2021,
  title = 	 {High-dimensional {Bayesian} optimization with sparse axis-aligned subspaces},
  author =       {Eriksson, David and Jankowiak, Martin},
  booktitle = 	 {Proceedings of the Thirty-Seventh Conference on Uncertainty in Artificial Intelligence},
  pages = 	 {493--503},
  year = 	 {2021},
  editor = 	 {de Campos, Cassio and Maathuis, Marloes H.},
  volume = 	 {161},
  series = 	 {Proceedings of Machine Learning Research},
  month = 	 {27--30 Jul},
  publisher =    {PMLR},
  url = 	 {https://proceedings.mlr.press/v161/eriksson21a.html}
}

@inproceedings{eriksson2019,
 author = {Eriksson, David and Pearce, Michael and Gardner, Jacob and Turner, Ryan D and Poloczek, Matthias},
 booktitle = {Advances in Neural Information Processing Systems},
 editor = {H. Wallach and H. Larochelle and A. Beygelzimer and F. d\textquotesingle Alch\'{e}-Buc and E. Fox and R. Garnett},
 pages = {},
 publisher = {Curran Associates, Inc.},
 title = {Scalable Global Optimization via Local Bayesian Optimization},
 url = {https://proceedings.neurips.cc/paper_files/paper/2019/file/6c990b7aca7bc7058f5e98ea909e924b-Paper.pdf},
 volume = {32},
 year = {2019}
}

@article{gomez2018,
author = {G{\'o}mez-Bombarelli, Rafael and Wei, Jennifer N. and Duvenaud, David and Hern{\'a}ndez-Lobato, Jos{\'e} Miguel and S{\'a}nchez-Lengeling, Benjam{\'\i}n and Sheberla, Dennis and Aguilera-Iparraguirre, Jorge and Hirzel, Timothy D. and Adams, Ryan P. and Aspuru-Guzik, Al{\'a}n},
title = {Automatic Chemical Design Using a Data-Driven Continuous Representation of Molecules},
journal = {ACS Central Science},
year = {2018},
volume = {4},
number = {2},
pages = {268--276},
publisher = {American Chemical Society},
doi = {10.1021/acscentsci.7b00572},
url = {https://doi.org/10.1021/acscentsci.7b00572},
issn = {2374-7943}
}

@article{hansen2001,
    author = {Hansen, Nikolaus and Ostermeier, Andreas},
    title = {Completely Derandomized Self-Adaptation in Evolution Strategies},
    journal = {Evolutionary Computation},
    volume = {9},
    number = {2},
    pages = {159-195},
    year = {2001},
    month = {06},
    issn = {1063-6560},
    doi = {10.1162/106365601750190398},
    url = {https://doi.org/10.1162/106365601750190398},
    eprint = {https://direct.mit.edu/evco/article-pdf/9/2/159/1493523/106365601750190398.pdf},
}

@inproceedings{jiao2023,
title={Crystal Structure Prediction by Joint Equivariant Diffusion on Lattices and Fractional Coordinates},
author={Rui Jiao and Wenbing Huang and Peijia Lin and Jiaqi Han and Pin Chen and Yutong Lu and Yang Liu},
booktitle={Workshop on ''Machine Learning for Materials'' ICLR 2023},
year={2023},
url={https://openreview.net/forum?id=VPByphdu24j}
}

@article{groeber2014,
author = {Groeber, Michael and Jackson, Michael},
year = {2014},
month = {02},
pages = {5},
title = {DREAM.3D: A Digital Representation Environment for the Analysis of Microstructure in 3D},
volume = {3},
journal = {Integrating Materials and Manufacturing Innovation},
doi = {10.1186/2193-9772-3-5}
}

@article{kulesza2012,
author = {Kulesza, Alex and Taskar, Ben},
year = {2012},
month = {07},
pages = {},
title = {Determinantal Point Processes for Machine Learning},
volume = {5},
journal = {Foundations and Trends® in Machine Learning},
doi = {10.1561/2200000044}
}

@article{markley2007,
author = {Markley, Landis and Cheng, Yang and Crassidis, John and Oshman, Yaakov},
year = {2007},
month = {07},
pages = {1193-1196},
title = {Averaging Quaternions},
volume = {30},
journal = {Journal of Guidance, Control, and Dynamics},
doi = {10.2514/1.28949}
}

@article{readshockley1950,
  title = {Dislocation Models of Crystal Grain Boundaries},
  author = {Read, W. T. and Shockley, W.},
  journal = {Phys. Rev.},
  volume = {78},
  issue = {3},
  pages = {275--289},
  numpages = {0},
  year = {1950},
  month = {May},
  publisher = {American Physical Society},
  doi = {10.1103/PhysRev.78.275},
  url = {https://link.aps.org/doi/10.1103/PhysRev.78.275}
}

@inproceedings{zhou2019,
  author={Zhou, Yi and Barnes, Connelly and Lu, Jingwan and Yang, Jimei and Li, Hao},
  booktitle={2019 IEEE/CVF Conference on Computer Vision and Pattern Recognition (CVPR)}, 
  title={On the Continuity of Rotation Representations in Neural Networks}, 
  year={2019},
  volume={},
  number={},
  pages={5738-5746},
  doi={10.1109/CVPR.2019.00589}}

@inproceedings{maus2022,
 author = {Maus, Natalie and Jones, Haydn and Moore, Juston and Kusner, Matt J and Bradshaw, John and Gardner, Jacob},
 booktitle = {Advances in Neural Information Processing Systems},
 editor = {S. Koyejo and S. Mohamed and A. Agarwal and D. Belgrave and K. Cho and A. Oh},
 pages = {34505--34518},
 publisher = {Curran Associates, Inc.},
 title = {Local Latent Space Bayesian Optimization over Structured Inputs},
 url = {https://proceedings.neurips.cc/paper_files/paper/2022/file/ded98d28f82342a39f371c013dfb3058-Paper-Conference.pdf},
 volume = {35},
 year = {2022}
}

@inproceedings{papenmeier2022,
title={Increasing the Scope as You Learn: Adaptive Bayesian Optimization in Nested Subspaces},
author={Leonard Papenmeier and Luigi Nardi and Matthias Poloczek},
booktitle={Advances in Neural Information Processing Systems},
editor={Alice H. Oh and Alekh Agarwal and Danielle Belgrave and Kyunghyun Cho},
year={2022},
url={https://openreview.net/forum?id=e4Wf6112DI}
}

@article{roters2019,
title = {DAMASK – The Düsseldorf Advanced Material Simulation Kit for modeling multi-physics crystal plasticity, thermal, and damage phenomena from the single crystal up to the component scale},
journal = {Computational Materials Science},
volume = {158},
pages = {420-478},
year = {2019},
issn = {0927-0256},
doi = {https://doi.org/10.1016/j.commatsci.2018.04.030},
url = {https://www.sciencedirect.com/science/article/pii/S0927025618302714},
author = {F. Roters and M. Diehl and P. Shanthraj and P. Eisenlohr and C. Reuber and S.L. Wong and T. Maiti and A. Ebrahimi and T. Hochrainer and H.-O. Fabritius and S. Nikolov and M. Friák and N. Fujita and N. Grilli and K.G.F. Janssens and N. Jia and P.J.J. Kok and D. Ma and F. Meier and E. Werner and M. Stricker and D. Weygand and D. Raabe}
}

@article{Wang2021,
  author  = {Wang, Cheng and Wang, Xiaogui and others},
  title   = {A comparative study of plastic deformation behaviors of {OFHC} copper based on crystal plasticity models},
  journal = {Journal of Materials Science},
  year    = {2021},
  volume  = {56},
  pages   = {8789--8814}
}

@article{ActaMg2014,
  author  = {Wang, F. and Sandl{\"o}bes, S. and Diehl, M. and Sharma, L. and Roters, F. and Raabe, D.},
  title   = {In situ observation of collective grain-scale mechanics in {Mg} and {Mg}--rare earth alloys},
  journal = {Acta Materialia},
  year    = {2014},
  volume  = {80},
  pages   = {77--93}
}

@article{FeCrAl2021,
  author  = {Zhang, Jingyu and Ding, Shurong and Du, Shiyu},
  title   = {A damage-effect-involved phenomenological crystal plasticity model and computational methods for mechanical responses of {FeCrAl} alloys},
  journal = {Materials Today Communications},
  year    = {2021},
  volume  = {28},
  pages   = {102595},
  doi     = {10.1016/j.mtcomm.2021.102595}
}

@inproceedings{stanton2022,
  title = 	 {Accelerating {B}ayesian Optimization for Biological Sequence Design with Denoising Autoencoders},
  author =       {Stanton, Samuel and Maddox, Wesley and Gruver, Nate and Maffettone, Phillip and Delaney, Emily and Greenside, Peyton and Wilson, Andrew Gordon},
  booktitle = 	 {Proceedings of the 39th International Conference on Machine Learning},
  pages = 	 {20459--20478},
  year = 	 {2022},
  editor = 	 {Chaudhuri, Kamalika and Jegelka, Stefanie and Song, Le and Szepesvari, Csaba and Niu, Gang and Sabato, Sivan},
  volume = 	 {162},
  series = 	 {Proceedings of Machine Learning Research},
  month = 	 {17--23 Jul},
  publisher =    {PMLR},
  url = 	 {https://proceedings.mlr.press/v162/stanton22a.html}
}

@inproceedings{tripp2020,
author = {Tripp, Austin and Daxberger, Erik and Hern\'{a}ndez-Lobato, Jos\'{e} Miguel},
title = {Sample-efficient optimization in the latent space of deep generative models via weighted retraining},
year = {2020},
isbn = {9781713829546},
publisher = {Curran Associates Inc.},
address = {Red Hook, NY, USA},
booktitle = {Proceedings of the 34th International Conference on Neural Information Processing Systems},
articleno = {945},
numpages = {14},
location = {Vancouver, BC, Canada},
series = {NIPS '20}
}

@article{wei2025,
author = {Wei, Ye and Peng, Bo and Xie, Ruiwen and Chen, Yangtao and Qin, Yu and Wen, Peng and Bauer, Stefan and Tung, Po-Yen and Raabe, Dierk},
title = {Deep active optimization for complex systems},
journal = {Nature Computational Science},
year = {2025},
volume = {5},
number = {9},
pages = {801--812},
doi = {10.1038/s43588-025-00858-x},
url = {https://doi.org/10.1038/s43588-025-00858-x},
issn = {2662-8457}
}

@inproceedings{xie2022,
title={Crystal Diffusion Variational Autoencoder for Periodic Material Generation},
author={Tian Xie and Xiang Fu and Octavian-Eugen Ganea and Regina Barzilay and Tommi S. Jaakkola},
booktitle={International Conference on Learning Representations},
year={2022},
url={https://openreview.net/forum?id=03RLpj-tc_}
}

@article{yang2018,
    author = {Yang, Zijiang and Li, Xiaolin and Catherine Brinson, L. and Choudhary, Alok N. and Chen, Wei and Agrawal, Ankit},
    title = {Microstructural Materials Design Via Deep Adversarial Learning Methodology},
    journal = {Journal of Mechanical Design},
    volume = {140},
    number = {11},
    pages = {111416},
    year = {2018},
    month = {10},
    issn = {1050-0472},
    doi = {10.1115/1.4041371},
    url = {https://doi.org/10.1115/1.4041371},
    eprint = {https://asmedigitalcollection.asme.org/mechanicaldesign/article-pdf/140/11/111416/6375275/md_140_11_111416.pdf},
}

@article{zeni2025,
author = {Zeni, Claudio and Pinsler, Robert and Z{\"u}gner, Daniel and Fowler, Andrew and Horton, Matthew and Fu, Xiang and Wang, Zilong and Shysheya, Aliaksandra and Crabb{\'e}, Jonathan and Ueda, Shoko and Sordillo, Roberto and Sun, Lixin and Smith, Jake and Nguyen, Bichlien and Schulz, Hannes and Lewis, Sarah and Huang, Chin-Wei and Lu, Ziheng and Zhou, Yichi and Yang, Han and Hao, Hongxia and Li, Jielan and Yang, Chunlei and Li, Wenjie and Tomioka, Ryota and Xie, Tian},
title = {A generative model for inorganic materials design},
journal = {Nature},
year = {2025},
volume = {639},
number = {8055},
pages = {624--632},
doi = {10.1038/s41586-025-08628-5},
url = {https://doi.org/10.1038/s41586-025-08628-5},
issn = {1476-4687}
}

@article{Yoo2023,
author = {Yoo, Pilsun and Bhowmik, Debsindhu and Mehta, Kshitij and Zhang, Pei and Liu, Frank and Lupo Pasini, Massimiliano and Irle, Stephan},
title = {Deep learning workflow for the inverse design of molecules with specific optoelectronic properties},
journal = {Scientific Reports},
year = {2023},
volume = {13},
number = {1},
pages = {20031},
doi = {10.1038/s41598-023-45385-9},
url = {https://doi.org/10.1038/s41598-023-45385-9},
issn = {2045-2322}
}

@article{Axelrod2022,
author = {Axelrod, Simon and G{\'o}mez-Bombarelli, Rafael},
title = {GEOM, energy-annotated molecular conformations for property prediction and molecular generation},
journal = {Scientific Data},
year = {2022},
volume = {9},
number = {1},
pages = {185},
doi = {10.1038/s41597-022-01288-4},
url = {https://doi.org/10.1038/s41597-022-01288-4},
issn = {2052-4463}
}

@article{Molesky2018,
author = {Molesky, Sean and Lin, Zin and Piggott, Alexander Y. and Jin, Weiliang and Vuckovi{\'c}, Jelena and Rodriguez, Alejandro W.},
title = {Inverse design in nanophotonics},
journal = {Nature Photonics},
year = {2018},
volume = {12},
number = {11},
pages = {659--670},
doi = {10.1038/s41566-018-0246-9},
url = {https://doi.org/10.1038/s41566-018-0246-9},
issn = {1749-4893}
}

@article{Zhou2021,
author = {Zhou, Ming and Liu, Dianjing and Belling, Samuel W. and Cheng, Haotian and Kats, Mikhail A. and Fan, Shanhui and Povinelli, Michelle L. and Yu, Zongfu},
title = {Inverse Design of Metasurfaces Based on Coupled-Mode Theory and Adjoint Optimization},
journal = {ACS Photonics},
year = {2021},
volume = {8},
number = {8},
pages = {2265--2273},
publisher = {American Chemical Society},
doi = {10.1021/acsphotonics.1c00100},
url = {https://doi.org/10.1021/acsphotonics.1c00100}
}

@misc{creswell2018,
      title={Inverting The Generator Of A Generative Adversarial Network (II)}, 
      author={Antonia Creswell and Anil A Bharath},
      year={2018},
      eprint={1802.05701},
      archivePrefix={arXiv},
      primaryClass={cs.CV},
      url={https://arxiv.org/abs/1802.05701}, 
}

@inproceedings{
li2026towards,
title={Towards Understanding the Mechanisms of Classifier-Free Guidance},
author={Xiang Li and Rongrong Wang and Qing Qu},
booktitle={The Thirty-ninth Annual Conference on Neural Information Processing Systems},
year={2026},
url={https://openreview.net/forum?id=bRAm7A02Qm}
}

@article{Liu2023,
author = {Liu, Han and Liu, Yuhan and Li, Kevin and Zhao, Zhangji and Schoenholz, Samuel S. and Cubuk, Ekin D. and Gupta, Puneet and Bauchy, Mathieu},
title = {End-to-end differentiability and tensor processing unit computing to accelerate materials' inverse design},
journal = {npj Computational Materials},
year = {2023},
volume = {9},
number = {1},
pages = {121},
doi = {10.1038/s41524-023-01080-x},
url = {https://doi.org/10.1038/s41524-023-01080-x},
issn = {2057-3960}
}

@article{Lu_2022,
    author = {Lu, Shuaihua and Zhou, Qionghua and Chen, Xinyu and Song, Zhilong and Wang, Jinlan},
    title = {Inverse design with deep generative models: next step in materials discovery},
    journal = {National Science Review},
    volume = {9},
    number = {8},
    pages = {nwac111},
    year = {2022},
    month = {08},
    issn = {2095-5138},
    doi = {10.1093/nsr/nwac111},
    url = {https://doi.org/10.1093/nsr/nwac111},
    eprint = {https://academic.oup.com/nsr/article-pdf/9/8/nwac111/45957070/nwac111.pdf},
}

@inbook{Frazier_2015,
   title={Bayesian Optimization for Materials Design},
   ISBN={9783319238715},
   ISSN={2196-2812},
   url={http://dx.doi.org/10.1007/978-3-319-23871-5_3},
   DOI={10.1007/978-3-319-23871-5_3},
   booktitle={Information Science for Materials Discovery and Design},
   publisher={Springer International Publishing},
   author={Frazier, Peter I. and Wang, Jialei},
   year={2015},
   month=Dec, pages={45–75} }

@Inbook{Abbott2016,
author="Abbott, T.
and Easton, M.
and Schmidt, R.",
editor="Mathaudhu, Suveen N.
and Luo, Alan A.
and Neelameggham, Neale R.
and Nyberg, Eric A.
and Sillekens, Wim H.",
title="Magnesium for Crashworthy Components",
bookTitle="Essential Readings in Magnesium Technology",
year="2016",
publisher="Springer International Publishing",
address="Cham",
pages="463--466",
isbn="978-3-319-48099-2",
doi="10.1007/978-3-319-48099-2_75",
url="https://doi.org/10.1007/978-3-319-48099-2_75"
}

@article{Chiu,
  title   = {Biodegradable magnesium alloys for orthopaedic applications},
  author  = {Lu, Yu and Deshmukh, Subodh and Jones, Ian and Chiu, Yu-Lung},
  journal = {Biomaterials Translation},
  volume  = {2},
  number  = {3},
  pages   = {214},
  doi     = {10.12336/biomatertransl.2021.03.005},
  year    = {2021}
}

@article{Bai2023,
title = {Applications of magnesium alloys for aerospace: A review},
journal = {Journal of Magnesium and Alloys},
volume = {11},
number = {10},
pages = {3609-3619},
year = {2023},
note = {Magnesium and Its Alloys for Better Future - JMA 10th Anniversary},
issn = {2213-9567},
doi = {https://doi.org/10.1016/j.jma.2023.09.015},
url = {https://www.sciencedirect.com/science/article/pii/S2213956723002281},
author = {Jingying Bai and Yan Yang and Chen Wen and Jing Chen and Gang Zhou and Bin Jiang and Xiaodong Peng and Fusheng Pan}
}

@article{Mason2009,
author = {Mason, J. K. and Schuh, C. A.},
title = {Expressing Crystallographic Textures through the Orientation Distribution Function: Conversion between Generalized Spherical Harmonic and Hyperspherical Harmonic Expansions},
journal = {Metallurgical and Materials Transactions A},
year = {2009},
volume = {40},
number = {11},
pages = {2590--2602},
doi = {10.1007/s11661-009-9936-8},
url = {https://doi.org/10.1007/s11661-009-9936-8},
issn = {1543-1940}
}

@article{Tuomo,
title = {Habit plane determination from reconstructed parent phase orientation maps},
journal = {Acta Materialia},
volume = {255},
pages = {119035},
year = {2023},
issn = {1359-6454},
doi = {https://doi.org/10.1016/j.actamat.2023.119035},
url = {https://www.sciencedirect.com/science/article/pii/S135964542300366X},
author = {Tuomo Nyyssönen and Azdiar A. Gazder and Ralf Hielscher and Frank Niessen}
}

@article{guru2025,
title = {Machine learning pipeline for Structure–Property modeling in Mg-alloys using microstructure and texture descriptors},
journal = {Acta Materialia},
volume = {295},
pages = {121132},
year = {2025},
issn = {1359-6454},
doi = {https://doi.org/10.1016/j.actamat.2025.121132},
url = {https://www.sciencedirect.com/science/article/pii/S1359645425004203},
author = {Mahish K. Guru and Jan Bohlen and Roland C. Aydin and Noomane Ben Khalifa}
}

@article{Xiong2016,
author = {Xiong, Wei and Olson, Gregory B.},
title = {Cybermaterials: materials by design and accelerated insertion of materials},
journal = {npj Computational Materials},
year = {2016},
volume = {2},
number = {1},
pages = {15009},
doi = {10.1038/npjcompumats.2015.9},
url = {https://doi.org/10.1038/npjcompumats.2015.9},
issn = {2057-3960}
}

@article{Yeung2023,
author = {Yeung, Christopher and Pham, Benjamin and Tsai, Ryan and Fountaine, Katherine T. and Raman, Aaswath P.},
title = {DeepAdjoint: An All-in-One Photonic Inverse Design Framework Integrating Data-Driven Machine Learning with Optimization Algorithms},
journal = {ACS Photonics},
year = {2023},
volume = {10},
number = {4},
pages = {884--891},
publisher = {American Chemical Society},
doi = {10.1021/acsphotonics.2c00968},
url = {https://doi.org/10.1021/acsphotonics.2c00968}
}

@misc{nie2020,
      title={TopologyGAN: Topology Optimization Using Generative Adversarial Networks Based on Physical Fields Over the Initial Domain}, 
      author={Zhenguo Nie and Tong Lin and Haoliang Jiang and Levent Burak Kara},
      year={2020},
      eprint={2003.04685},
      archivePrefix={arXiv},
      primaryClass={cs.CE},
      url={https://arxiv.org/abs/2003.04685}, 
}

@inproceedings{
zeng2024antibody,
title={Antibody Design with Constrained Bayesian Optimization},
author={Yimeng Zeng and Hunter Elliott and Phillip Maffettone and Peyton Greenside and Osbert Bastani and Jacob R. Gardner},
booktitle={ICLR 2024 Workshop on Generative and Experimental Perspectives for Biomolecular Design},
year={2024},
url={https://openreview.net/forum?id=K5Sr6WSA4B}
}

@article{KusampudiDiehl2023,
title = {Inverse design of dual-phase steel microstructures using generative machine learning model and Bayesian optimization},
journal = {International Journal of Plasticity},
volume = {171},
pages = {103776},
year = {2023},
issn = {0749-6419},
doi = {https://doi.org/10.1016/j.ijplas.2023.103776},
url = {https://www.sciencedirect.com/science/article/pii/S0749641923002607},
author = {Navyanth Kusampudi and Martin Diehl}
}

@article{ParkKushwaha2024,
title = {Nonlinear inverse design of mechanical multi-material metamaterials enabled by video denoising diffusion and structure identifier},
journal = {Engineering Applications of Artificial Intelligence},
volume = {172},
pages = {114368},
year = {2026},
issn = {0952-1976},
doi = {https://doi.org/10.1016/j.engappai.2026.114368},
url = {https://www.sciencedirect.com/science/article/pii/S0952197626006494},
author = {Jaewan Park and Shashank Kushwaha and Junyan He and Seid Koric and Qibang Liu and Iwona Jasiuk and Diab Abueidda}
}

@article{Zheng2026,
author = {Zheng, Li and Kumar, Siddhant and Kochmann, Dennis M.},
title = {Algebraic language models for inverse design of metamaterials via diffusion transformers},
journal = {Nature Machine Intelligence},
year = {2026},
volume = {8},
number = {4},
pages = {628--640},
doi = {10.1038/s42256-026-01218-8},
url = {https://doi.org/10.1038/s42256-026-01218-8},
issn = {2522-5839}
}

@article{Rassloff2026,
author = {Ra{\ss}loff, Alexander and Seibert, Paul and Kalina, Karl A. and K{\"a}stner, Markus},
title = {Inverse design of spinodoid structures using Bayesian optimization},
journal = {Computational Mechanics},
year = {2026},
volume = {77},
number = {1},
pages = {275--296},
doi = {10.1007/s00466-024-02587-w},
url = {https://doi.org/10.1007/s00466-024-02587-w},
issn = {1432-0924}
}

@inproceedings{radford2021clip,
  title     = {Learning Transferable Visual Models From Natural Language Supervision},
  author    = {Alec Radford and Jong Wook Kim and Chris Hallacy and Aditya Ramesh and Gabriel Goh and Sandhini Agarwal and Girish Sastry and Amanda Askell and Pamela Mishkin and Jack Clark and Gretchen Krueger and Ilya Sutskever},
  booktitle = {Proceedings of the 38th International Conference on Machine Learning},
  editor    = {Marina Meila and Tong Zhang},
  series    = {Proceedings of Machine Learning Research},
  volume    = {139},
  pages     = {8748--8763},
  year      = {2021},
  publisher = {PMLR},
  url       = {https://proceedings.mlr.press/v139/radford21a.html}
}

@inproceedings{schuhmann2022laion5b,
  title     = {{LAION-5B}: An open large-scale dataset for training next generation image-text models},
  author    = {Christoph Schuhmann and Romain Beaumont and Richard Vencu and Cade Gordon and Ross Wightman and Mehdi Cherti and Theo Coombes and Aarush Katta and Clayton Mullis and Mitchell Wortsman and Patrick Schramowski and Srivatsa Kundurthy and Katherine Crowson and Ludwig Schmidt and Robert Kaczmarczyk and Jenia Jitsev},
  booktitle = {Advances in Neural Information Processing Systems},
  volume    = {35},
  pages     = {25278--25294},
  year      = {2022},
  publisher = {Curran Associates, Inc.},
  url       = {https://papers.nips.cc/paper_files/paper/2022/hash/a1859debfb3b59d094f3504d5ebb6c25-Abstract-Datasets_and_Benchmarks.html}
}

@article{podell2023sdxl,
  title   = {{SDXL}: Improving Latent Diffusion Models for High-Resolution Image Synthesis},
  author  = {Dustin Podell and Zion English and Kyle Lacey and Andreas Blattmann and Tim Dockhorn and Jonas M{\"u}ller and Joe Penna and Robin Rombach},
  journal = {arXiv preprint arXiv:2307.01952},
  year    = {2023},
  url     = {https://arxiv.org/abs/2307.01952}
}

@article{esser2024sd3,
  title   = {Scaling Rectified Flow Transformers for High-Resolution Image Synthesis},
  author  = {Patrick Esser and Sumith Kulal and Andreas Blattmann and Rahim Entezari and Jonas M{\"u}ller and Harry Saini and Yam Levi and Dominik Lorenz and Axel Sauer and Frederic Boesel and Dustin Podell and Tim Dockhorn and Zion English and Kyle Lacey and Alex Goodwin and Yannik Marek and Robin Rombach},
  journal = {arXiv preprint arXiv:2403.03206},
  year    = {2024},
  url     = {https://arxiv.org/abs/2403.03206}
}

@inproceedings{peebles2023dit,
  author    = {William Peebles and Saining Xie},
  title     = {Scalable Diffusion Models with Transformers},
  booktitle = {Proceedings of the IEEE/CVF International Conference on Computer Vision (ICCV)},
  month     = {October},
  year      = {2023},
  pages     = {4195--4205},
  url       = {https://openaccess.thecvf.com/content/ICCV2023/html/Peebles_Scalable_Diffusion_Models_with_Transformers_ICCV_2023_paper.html}
}

@inproceedings{esser2021vqgan,
  author    = {Patrick Esser and Robin Rombach and Bj{\"o}rn Ommer},
  title     = {Taming Transformers for High-Resolution Image Synthesis},
  booktitle = {Proceedings of the IEEE/CVF Conference on Computer Vision and Pattern Recognition (CVPR)},
  month     = {June},
  year      = {2021},
  pages     = {12873--12883},
  url       = {https://openaccess.thecvf.com/content/CVPR2021/html/Esser_Taming_Transformers_for_High-Resolution_Image_Synthesis_CVPR_2021_paper.html}
}

@inproceedings{lipman2023flowmatching,
  title     = {Flow Matching for Generative Modeling},
  author    = {Yaron Lipman and Ricky T. Q. Chen and Heli Ben-Hamu and Maximilian Nickel and Matt Le},
  booktitle = {Proceedings of the International Conference on Learning Representations},
  year      = {2023},
  url       = {https://openreview.net/forum?id=PqvMRDCJT9t}
}

@article{ho2022cfg,
  title   = {Classifier-Free Diffusion Guidance},
  author  = {Jonathan Ho and Tim Salimans},
  journal = {arXiv preprint arXiv:2207.12598},
  year    = {2022},
  url     = {https://arxiv.org/abs/2207.12598}
}

@inproceedings{bardes2022vicreg,
  title     = {{VICReg}: Variance-Invariance-Covariance Regularization for Self-Supervised Learning},
  author    = {Adrien Bardes and Jean Ponce and Yann LeCun},
  booktitle = {Proceedings of the International Conference on Learning Representations},
  year      = {2022},
  url       = {https://openreview.net/forum?id=xm6YD62D1Ub}
}

@inproceedings{heusel2017gans,
  title     = {GANs Trained by a Two Time-Scale Update Rule Converge to a Local Nash Equilibrium},
  author    = {Heusel, Martin and Ramsauer, Hubert and Unterthiner, Thomas and Nessler, Bernhard and Hochreiter, Sepp},
  booktitle = {Advances in Neural Information Processing Systems},
  volume    = {30},
  pages     = {6626--6637},
  year      = {2017}
}

@article{Wang2003MultiscaleSS,
  title   = {Multiscale structural similarity for image quality assessment},
  author  = {Zhou Wang and Eero P. Simoncelli and Alan Conrad Bovik},
  journal = {The Thrity-Seventh Asilomar Conference on Signals, Systems \& Computers, 2003},
  year    = {2003},
  volume  = {2},
  pages   = {1398--1402 Vol.2},
  url     = {https://api.semanticscholar.org/CorpusID:60600316}
}

@inproceedings{zhang2018perceptual,
  title     = {The Unreasonable Effectiveness of Deep Features as a Perceptual Metric},
  author    = {Zhang, Richard and Isola, Phillip and Efros, Alexei A and Shechtman, Eli and Wang, Oliver},
  booktitle = {CVPR},
  year      = {2018}
}

@inproceedings{krishnamoorthy2023ddom,
  title = 	 {Diffusion Models for Black-Box Optimization},
  author =       {Krishnamoorthy, Siddarth and Mashkaria, Satvik Mehul and Grover, Aditya},
  booktitle = 	 {Proceedings of the 40th International Conference on Machine Learning},
  pages = 	 {17842--17857},
  year = 	 {2023},
  editor = 	 {Krause, Andreas and Brunskill, Emma and Cho, Kyunghyun and Engelhardt, Barbara and Sabato, Sivan and Scarlett, Jonathan},
  volume = 	 {202},
  series = 	 {Proceedings of Machine Learning Research},
  month = 	 {23--29 Jul},
  publisher =    {PMLR},
  url = 	 {https://proceedings.mlr.press/v202/krishnamoorthy23a.html}
}

@inproceedings{black2024ddpo,
title={Training Diffusion Models with Reinforcement Learning},
author={Kevin Black and Michael Janner and Yilun Du and Ilya Kostrikov and Sergey Levine},
booktitle={ICML 2023 Workshop The Many Facets of Preference-Based Learning},
year={2023},
url={https://openreview.net/forum?id=w5QmbPyzfg}
}

@inproceedings{uehara2024seiko,
  title = 	 {Feedback Efficient Online Fine-Tuning of Diffusion Models},
  author =       {Uehara, Masatoshi and Zhao, Yulai and Black, Kevin and Hajiramezanali, Ehsan and Scalia, Gabriele and Diamant, Nathaniel Lee and Tseng, Alex M and Levine, Sergey and Biancalani, Tommaso},
  booktitle = 	 {Proceedings of the 41st International Conference on Machine Learning},
  pages = 	 {48892--48918},
  year = 	 {2024},
  editor = 	 {Salakhutdinov, Ruslan and Kolter, Zico and Heller, Katherine and Weller, Adrian and Oliver, Nuria and Scarlett, Jonathan and Berkenkamp, Felix},
  volume = 	 {235},
  series = 	 {Proceedings of Machine Learning Research},
  month = 	 {21--27 Jul},
  publisher =    {PMLR},
  url = 	 {https://proceedings.mlr.press/v235/uehara24a.html}
}

@article{karniadakis2021piml,
  title   = {Physics-informed machine learning},
  author  = {Karniadakis, George Em and Kevrekidis, Ioannis G. and Lu, Lu and Perdikaris, Paris and Wang, Sifan and Yang, Liu},
  journal = {Nature Reviews Physics},
  volume  = {3},
  number  = {6},
  pages   = {422--440},
  year    = {2021},
  doi     = {10.1038/s42254-021-00314-5}
}

@article{vonrueden2021informed,
  author={von Rueden, Laura and Mayer, Sebastian and Beckh, Katharina and Georgiev, Bogdan and Giesselbach, Sven and Heese, Raoul and Kirsch, Birgit and Pfrommer, Julius and Pick, Annika and Ramamurthy, Rajkumar and Walczak, Michal and Garcke, Jochen and Bauckhage, Christian and Schuecker, Jannis},
  journal={IEEE Transactions on Knowledge and Data Engineering}, 
  title={Informed Machine Learning – A Taxonomy and Survey of Integrating Prior Knowledge into Learning Systems}, 
  year={2023},
  volume={35},
  number={1},
  pages={614-633},
  doi={10.1109/TKDE.2021.3079836}}

@article{bostanabad2018,
title = {Computational microstructure characterization and reconstruction: Review of the state-of-the-art techniques},
journal = {Progress in Materials Science},
volume = {95},
pages = {1-41},
year = {2018},
issn = {0079-6425},
doi = {https://doi.org/10.1016/j.pmatsci.2018.01.005},
url = {https://www.sciencedirect.com/science/article/pii/S0079642518300112},
author = {Ramin Bostanabad and Yichi Zhang and Xiaolin Li and Tucker Kearney and L. Catherine Brinson and Daniel W. Apley and Wing Kam Liu and Wei Chen}
}

@article{generale2024,
title = {Inverse stochastic microstructure design},
journal = {Acta Materialia},
volume = {271},
pages = {119877},
year = {2024},
issn = {1359-6454},
doi = {https://doi.org/10.1016/j.actamat.2024.119877},
url = {https://www.sciencedirect.com/science/article/pii/S1359645424002301},
author = {Adam P. Generale and Andreas E. Robertson and Conlain Kelly and Surya R. Kalidindi}
}

@article{buzzy2025,
title = {Active learning for the design of polycrystalline textures using conditional normalizing flows},
journal = {Acta Materialia},
volume = {284},
pages = {120537},
year = {2025},
issn = {1359-6454},
doi = {https://doi.org/10.1016/j.actamat.2024.120537},
url = {https://www.sciencedirect.com/science/article/pii/S1359645424008863},
author = {Michael O. Buzzy and David {Montes de Oca Zapiain} and Adam P. Generale and Surya R. Kalidindi and Hojun Lim}
}

@inproceedings{wu2024cindm,
title={Compositional Generative Inverse Design},
author={Tailin Wu and Takashi Maruyama and Long Wei and Tao Zhang and Yilun Du and Gianluca Iaccarino and Jure Leskovec},
booktitle={The Twelfth International Conference on Learning Representations},
year={2024},
url={https://openreview.net/forum?id=wmX0CqFSd7}
}

@inproceedings{bastek2025pidm,
title={Physics-Informed Diffusion Models},
author={Jan-Hendrik Bastek and WaiChing Sun and Dennis Kochmann},
booktitle={The Thirteenth International Conference on Learning Representations},
year={2025},
url={https://openreview.net/forum?id=tpYeermigp}
}
}

\appendix

\section{Extended E2 results}
\label{app:extended-results}

This appendix reports the toughness-weighted objective results that complement the target-driven E2 results in Sec.~\ref{sec:e2}. The protocol is unchanged: the shared seed cache is used only for warm start, all curves and tables exclude that cache, and each run receives $160$ optimizer-phase simulator evaluations. Average wall-clock runtime for these $J_{V1}$ runs was $8.48$ hours on an H200 machine, corresponding to an estimated $5.94$ kWh of GPU energy per run under the same $700$ W TDP accounting used in the main text.

\begin{figure}[H]
  \centering
  \includegraphics[width=\linewidth]{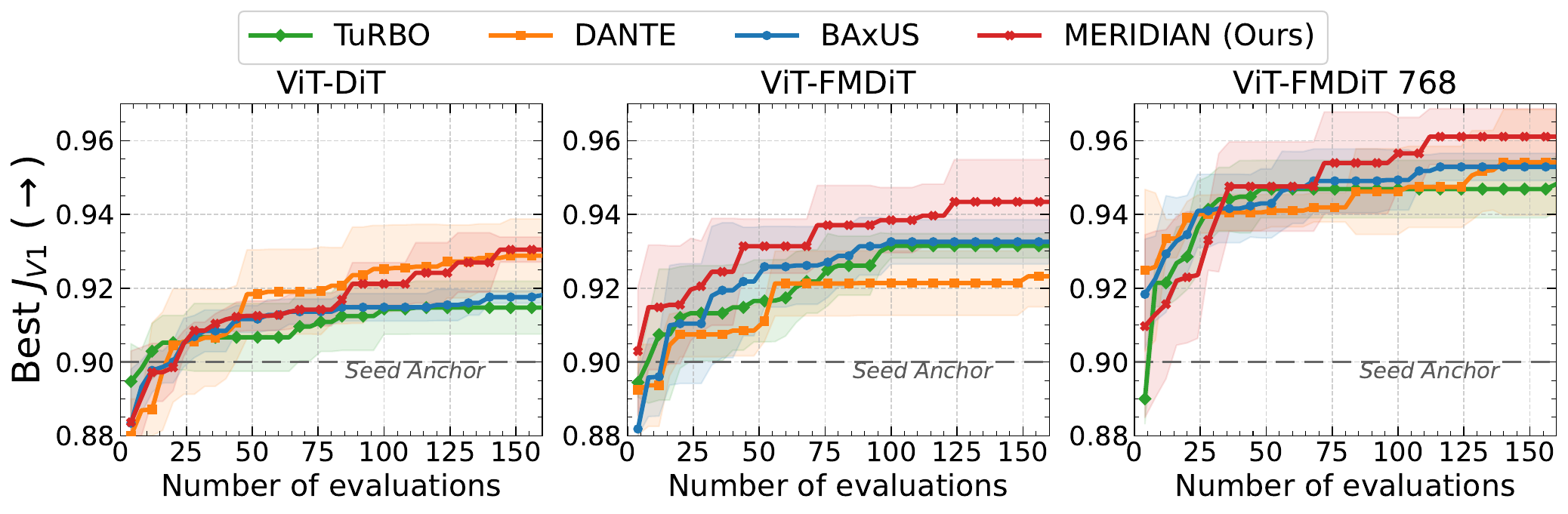}
  \caption{\textbf{Best Mean Toughness-weighted $J_{V1}$} versus num. of evaluations (batch size 4, $n_{\mathrm{seed}}= 5$).}
  \label{fig:e2-v1-convergence}
\end{figure}

\begin{figure}[!ht]
  \begin{minipage}[c]{0.3\textwidth}
    \centering
    \includegraphics[width=0.9\linewidth]{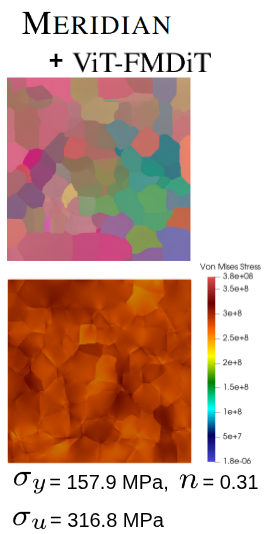}
    \captionof{figure}{Best $J_{V1}$ \\ microstructure and stress field.}
    \label{fig:placeholder-v1}
  \end{minipage}%
  \hfill
  \begin{minipage}[c]{0.7\textwidth}
    \centering
    \captionof{table}{\textbf{Optimization Results:} Toughness-Weighted, $J_{V1}$ ($n_{\mathrm{seed}}= 5$).}
    \label{tab:opt-v1}
    \footnotesize
    \setlength{\tabcolsep}{3pt}
    \begin{tabular}{llccccc}
      \toprule
      Decoder & Optimizer & $J_{V1}\!\uparrow$ & $\sigy$ & $n$ & $\sigu$ & \# sims \\
      \midrule
      \rowcolor{gray!10} \cellcolor{white}\multirow{4}{*}{ViT{-}DiT}
        & \textbf{\meridian{}}  & $0.930 \pm 0.003^*$ & $154.3$ & $0.308$ & $314.0$ & $144$ \\
        & \dante{}               & $0.929 \pm 0.010$ & $155.6$ & $0.305$ & $313.4$ & $148$ \\
        & \turbo{}               & $0.915 \pm 0.007$ & $153.8$ & $0.306$ & $311.3$ & $112$ \\
        & \baxus{}               & $0.918 \pm 0.002$ & $151.8$ & $0.312$ & $314.2$ & $160$ \\
      \midrule
      \rowcolor{gray!10} \cellcolor{white}\multirow{4}{*}{ViT{-}FMDiT}
        & \textbf{\meridian{}}  & $0.943 \pm 0.011^*$ & $156.8$ & $0.306$ & $316.1$ & $124$ \\
        & \dante{}               & $0.923 \pm 0.008$ & $155.2$ & $0.305$ & $313.0$ & $152$ \\
        & \turbo{}               & $0.931 \pm 0.003$ & $154.0$ & $0.309$ & $314.2$ & $100$ \\
        & \baxus{}               & $0.933 \pm 0.006$ & $156.3$ & $0.301$ & $311.7$ & $100$ \\
      \midrule
      \rowcolor{blue!8} \cellcolor{white}\multirow{4}{*}{\shortstack{ViT{-}FMDiT\\[1pt]{\footnotesize 768}}}
        & \textbf{\meridian{}}$^\dagger$  & $\mathbf{0.961 \pm 0.008}^*$ & $157.9$ & $0.305$ & $316.8$ & $112$ \\
        & \dante{}               & $0.954 \pm 0.014$ & $159.8$ & $0.300$ & $316.6$ & $140$ \\
        & \turbo{}               & $0.948 \pm 0.007$ & $156.1$ & $0.309$ & $317.9$ & $160$ \\
        & \baxus{}               & $0.953 \pm 0.004$ & $157.2$ & $0.304$ & $315.1$ & $116$ \\
      \midrule
      \rowcolor{gray!10} \cellcolor{white}\multirow{4}{*}{\shortstack{ViT{-}FMDiT\\[1pt]{\footnotesize 1024}}}
        & \textbf{\meridian{}}  & $0.952 \pm 0.005^*$ & $156.2$ & $0.308$ & $317.6$ & $160$ \\
        & \dante{}               & $0.923 \pm 0.006$ & $153.5$ & $0.309$ & $314.0$ & $156$ \\
        & \turbo{}               & $0.944 \pm 0.010$ & $157.2$ & $0.305$ & $315.3$ & $140$ \\
        & \baxus{}               & $0.936 \pm 0.011$ & $155.8$ & $0.308$ & $316.7$ & $128$ \\
      \bottomrule
      \multicolumn{7}{l}{\scriptsize \colorbox{gray!10}{Grey} = our method (\meridian{}). \; $^*$ = best per decoder block. \; \colorbox{blue!8}{$^\dagger$} = best overall.}
    \end{tabular}
    \vspace{1pt}
  \end{minipage}
\end{figure}

The objective changes the design question from hitting a prescribed stress--strain tuple to finding high-strength, high-plastic-work microstructures without dropping below the hardening floor. Figure~\ref{fig:e2-v1-convergence} and Table~\ref{tab:opt-v1} show that the same decoder--optimizer pairing remains the most effective overall: ViT{-}FMDiT-$768$ with \meridian{} reaches $J_{V1}=0.961\pm0.008$, with $\sigy=157.9$ MPa, $n=0.305$, $\sigu=316.8$ MPa, and $112$ optimizer-phase simulations to the best value. Compared with the target-driven case, the best solution now moves slightly toward higher ultimate strength while preserving hardening, as expected for a toughness-weighted score.

The $1024$-D appendix rows mirror the main-text interpretation. This does not indicate a defective decoder; rather, the wider bottleneck exposes isolated texture pockets that are difficult to model globally from $160$ calls. Sparse local perturbations and low-dimensional PCA-aligned subspace moves can exploit such pockets without fitting the whole ambient latent at once, which explains why the best $J_{V1}$ rows for ViT{-}FMDiT-$1024$ come from those search biases. Across both objectives, the consistent conclusion is that ViT{-}FMDiT-$768$ gives the best operating point for \Co{}: enough latent capacity to express useful AZ31 textures, but not so much that the expensive-oracle optimizer loses sample efficiency.

\section{Additional ablations and sensitivity of \meridian{}}
\label{app:ablations}

\textbf{Ranking consistency.} Our conclusions are ranking statements (which optimizer and which decoder are best), so we test whether the rankings survive variations in the control parameters. Across four decoders and two objectives (Tabs.~\ref{tab:opt-v2},~\ref{tab:opt-v1}), \meridian{} attains the best mean objective in all eight studies; under a null of uniformly random ranking among the four optimizers run on every decoder, this occurs with probability $(1/4)^8 \approx 1.5\times10^{-5}$. Symmetrically, ViT-FMDiT-$768$ is the best decoder in all eight optimizer--objective combinations. The ViT-DiT decoder was trained under a materially different loss-weight configuration from the ViT-FMDiT decoders ($\lambda_{\mathrm{vic}}$ $0.01$ vs.\ $0.05$, $\lambda_{\mathrm{freq}}$ $0.1$ vs.\ $0.3$, $\lambda_{\mathrm{con}}$ starting at $0.0$ vs.\ $0.3$, different onset/ramp epochs), so the optimizer ranking is not an artifact of one loss-weight specification.

\textbf{DPP versus greedy top-$k$ batch selection.} We swap only \meridian{}'s DPP batch selector for plain top-$k$ by acquisition score, holding every other component and the seed cache fixed ($J_{V2}$, $160$ evaluations, matched seeds; Tab.~\ref{tab:dpp-topk}). These are separate matched runs, so values differ slightly from Tab.~\ref{tab:opt-v2}. On ViT-FMDiT-$768$ both rules reach the same mean quality ($J_{V2}=-0.138$), so DPP does not raise the ceiling, but it reaches the plateau about $11\%$ of the budget earlier ($143$ vs.\ $160$ evaluations) with lower run-to-run spread ($\pm0.007$ vs.\ $\pm0.010$), and it eliminates a batch-collapse failure mode in which one top-$k$ seed selected four near-duplicate candidates and made no progress after the first iteration. On ViT-FMDiT-$1024$ the two rules tie statistically ($-0.165\pm0.010$ vs.\ $-0.162\pm0.006$, no collapsed seeds). Under an expensive oracle and a thin-shell latent manifold, batch diversity therefore buys sample efficiency and robustness rather than peak performance.

\begin{table}[!ht]
  \caption{\textbf{Batch-rule ablation} ($J_{V2}$, $160$ evaluations). Only the batch selector differs; ``collapsed'' counts seeds in which the batch degenerated to near-duplicates.}
  \label{tab:dpp-topk}
  \centering
  \small
  \setlength{\tabcolsep}{3pt}
  \begin{tabular}{llcccccc}
    \toprule
    Decoder & Batch rule & $J_{V2}\uparrow$ & $\sigy$ (MPa) & $n$ & $\sigu$ (MPa) & evals$\to$best & collapsed \\
    \midrule
    \multirow{2}{*}{ViT-FMDiT-768}  & \textbf{DPP (ours)} & $\mathbf{-0.138 \pm 0.007}$ & $157.5$ & $0.303$ & $315.0$ & $\mathbf{143}$ & $\mathbf{0/3}$ \\
                                    & top-$k$             & $-0.138 \pm 0.010$ & $157.6$ & $0.303$ & $314.7$ & $160$ & $1/3$ \\
    \midrule
    \multirow{2}{*}{ViT-FMDiT-1024} & \textbf{DPP (ours)} & $-0.165 \pm 0.010$ & $153.6$ & $0.305$ & $309.7$ & $153$ & $0/4$ \\
                                    & top-$k$             & $-0.162 \pm 0.006$ & $154.4$ & $0.306$ & $312.3$ & $146$ & $0/4$ \\
    \bottomrule
  \end{tabular}
\end{table}

\textbf{Sensitivity to \meridian{}'s control parameters.} \meridian{} has its own settings: trust-region size and shrink rate, the latent-norm shell bound $k_\sigma$, and the batch-diversity term. We grouped repeated ViT-FMDiT-$768$ \meridian{} runs under one identical pair of objectives by the configuration they were produced under (Tab.~\ref{tab:meridian-sensitivity}). The evaluation budget is essentially unaffected: under $J_{V2}$ every setting reaches its plateau at $104$--$106$ of the $160$ evaluations, and no setting stalls, diverges, or fails to converge. The achieved optimum moves by about one seed-level standard deviation ($0.011$ on $J_{V2}$, against a per-setting seed spread of $0.005$--$0.013$). The published setting is best on both objectives although it was tuned only on $J_{V1}$. The one setting with a systematic effect is the batch-diversity term, a claimed component of the method ablated above.

\begin{table}[!ht]
  \caption{\textbf{\meridian{} control-parameter sensitivity} on ViT-FMDiT-$768$. ``evals'' = evaluations to plateau.}
  \label{tab:meridian-sensitivity}
  \centering
  \small
  \setlength{\tabcolsep}{4pt}
  \begin{tabularx}{\linewidth}{@{}l X cc cc@{}}
    \toprule
    \meridian{} setting & What differs & $J_{V1}\uparrow$ & evals & $J_{V2}\uparrow$ & evals \\
    \midrule
    published reference (Alg.~\ref{alg:meridian}) & --- & $\mathbf{+0.961 \pm 0.009}$ & $58$ & $\mathbf{-0.136 \pm 0.009}$ & $106$ \\
    earlier trust-region + shell & compact/fixed trust region, DPP pool $64$, shell $k_\sigma\in\{1.0,2.0\}$, faster/slower shrink & $+0.944 \pm 0.003$ & $53$ & $-0.147 \pm 0.005$ & $104$ \\
    DPP $\to$ greedy top-$k$ & diversity term removed & $+0.956 \pm 0.012$ & $93$ & $-0.142 \pm 0.013$ & $105$ \\
    \midrule
    \textbf{spread (max$-$min)} & & $\mathbf{0.017}$ & & $\mathbf{0.011}$ & \\
    \bottomrule
  \end{tabularx}
\end{table}

\section{Encoder--decoder details}
\label{app:enc-dec}

This appendix records the implementation detail behind the encoder--decoder family that the main text only sketches. The goal is not to restate Sec.~\ref{sec:enc-dec}, but to make the exact architectural, training, and validation choices reproducible: the shared encoder, the three conditioning interfaces, the composite training objective, and the bottleneck-capacity ablation.

\textbf{Shared encoder --- exact specification}. The encoder is OpenCLIP ViT-H/14 (\texttt{laion2b\_s32b\_b79k} weights)~\citep{radford2021clip,schuhmann2022laion5b}: $32$ transformer blocks, hidden width $1280$, patch size $14$, and input resolution $512{\times}512$, which yields $37{\times}37{=}1369$ patch tokens after bicubic interpolation of the original $16{\times}16$ positional embedding. We keep CLIP normalisation throughout rather than switching to ImageNet statistics. The freezing pattern is deliberately asymmetric: \texttt{conv1}, \texttt{cls\_token}, the positional embedding, \texttt{ln\_pre}, and the lower $16$ transformer blocks remain frozen, while only the upper $16$ blocks are trainable, with gradient checkpointing enabled there. This keeps most of the LAION prior intact while allowing the top of the trunk to adapt to EBSD imagery. The CLS readout is replaced by a multi-query attention pooler with $4$ learned queries (initialised $\mathcal{N}(0,0.02)$), $8$ attention heads, query self-attention, and a per-query FFN, followed by a merge head $\mathrm{Linear}(4D \to D)$. The bottleneck MLP is $\mathrm{Linear}(1280,1280)\to\mathrm{GELU}\to\mathrm{Linear}(1280,d)\to\mathrm{LayerNorm}$ with $d \in \{512,768,1024\}$.

\textbf{Conditioning adaptors --- exact specifications}. All three decoders reuse the same bottleneck but expose different native conditioning interfaces, so each adaptor is designed to match the backbone it plugs into rather than to force a uniform abstraction.
\textbf{FM-DiT.} $\textsc{LatentToTokens}(d \to 16{\times}4096)$ uses $16$ learned queries, $5$ AdaLN-Zero blocks, and QK-normalised cross-attention against the pooled latent. Its outer $\mathrm{proj\_out}$ is zero-initialised, which is stable here because flow matching learns a residual velocity field. The pooled head is $\mathrm{Linear}(d,2048)\to\mathrm{SiLU}\to\mathrm{Linear}(2048,2048)$, again with a zero-initialised last layer. The adaptor carries roughly $11$M trainable parameters and the pooled head another $4$M. The frozen backbone is \texttt{stabilityai/stable-diffusion-3.5-medium}; its VAE uses $16$ channels, $8{\times}$ downsampling, and scale factor $1.5305$. Timesteps are sampled as $t = \sigma(\mathcal{N}(0,1)+\log\,\mathrm{shift})$ with $\mathrm{shift}{=}3.0$, matching the published SD3.5 logit-normal schedule, and inference uses FlowMatchEulerDiscrete.
\textbf{SDXL.} $\textsc{LatentToTokens}(d \to 77{\times}2048)$ mirrors the original SDXL text sequence length with $77$ learned queries passed through $3$ AdaLN-Zero blocks. Here zero-initialisation was empirically unstable, so both the outer projection and the pooled head are initialised with Xavier gain $0.10$; runs with zero init produced dead conditioning heads. The pooled branch is $\mathrm{Linear}(d,1280)\to\mathrm{SiLU}\to\mathrm{Linear}(1280,1280)$ feeding \texttt{add\_embedding}, and a learned $\textsc{NullToken}$ implements classifier-free guidance dropout with $p_{\mathrm{drop}}{=}0.1$. The trainable interface is roughly $8$M parameters in the token adaptor, $0.8$M in the pooled head, plus the $512$-parameter null token. The SDXL UNet backbone remains frozen in bf16; training uses DDPM and inference EulerDiscrete.
\textbf{DiT.} DiT-XL/2 (hidden size $1152$, $28$ blocks, $16$ heads, patch size $2$, VAE \texttt{sd-vae-ft-mse}) is fully trainable because freezing failed under the $256{\to}512$ transfer. The class-conditioning slot is repurposed through $\mathrm{ClassProj}(d \to 1152) = \mathrm{Linear}\to\mathrm{GELU}\to\mathrm{Linear}$ with Xavier initialisation. A pair of forward pre-hooks (\texttt{\_CLIPInjector}/\texttt{\_ZeroEmbedder}) removes the original repeated class-embedding path and replaces it with a single top-of-stack injection of $\mathrm{ClassProj}(z)$. In parallel, $\textsc{LatentToSpatialTokens}(d \to 16{\times}1152)$ emits $16$ spatial tokens that a $\textsc{SpatialCrossFuser}$ inserts, with zero-initialised output projection, at blocks $\{0,7,14,21\}$. Training uses DDPM with $1000$ steps and linearly spaced $\beta$ from $10^{-4}$ to $0.02$; inference uses $250$-step DDIM.

\begin{figure}[!ht]
  \centering
  \includegraphics[width=\linewidth]{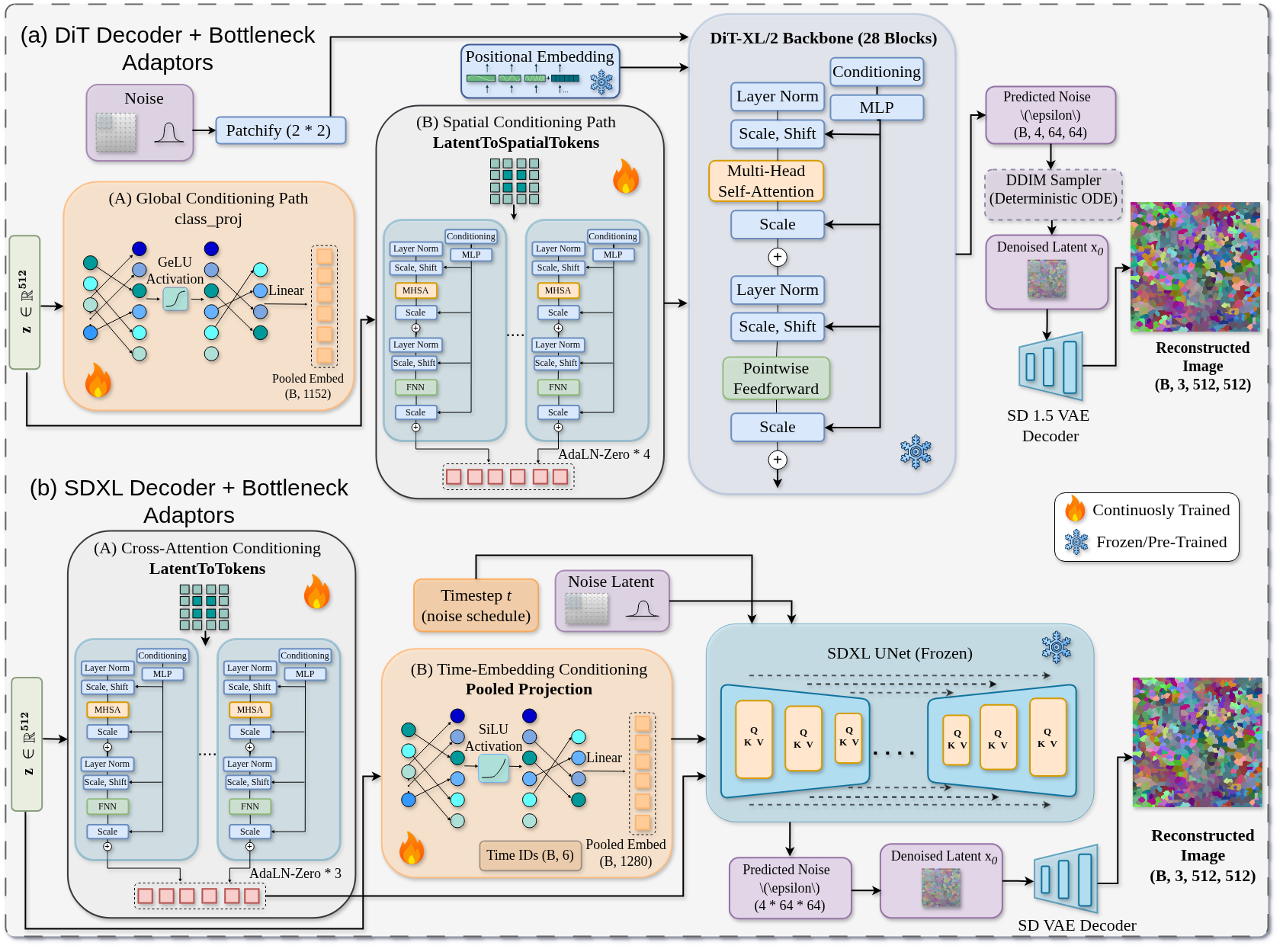}
  \caption{\textbf{Companion decoder pipelines} (deferred from Fig.~\ref{fig:archs}). (a)~ViT$+$DiT-XL/2 (\emph{trainable}, $256{\to}512$): $\textsc{ClassProj}$ repurposes the DiT class-embedding slot to inject $z$; $16{\times}1152$ spatial tokens; SD~1.5 VAE. (b)~ViT$+$SDXL (frozen): $77{\times}2048$ tokens via $3$ AdaLN-Zero blocks (Xavier-$0.10$) plus $\textsc{Pooled Projection}$ $\to$ $1280$-D; SD-XL VAE.}
  \label{fig:archs-appendix}
\end{figure}

\textbf{Composite loss and training schedule}. Training combines the five terms introduced in Eq.~\eqref{eq:composite-loss} rather than relying on the backbone loss alone. The diffusion term $\mathcal{L}_{\mathrm{diff}}$ is the rectified-flow velocity MSE $\lVert v_\theta(x_t,t,z) - (\epsilon - x_0)\rVert_2^2$ for ViT-FMDiT and the standard $\epsilon$-prediction MSE $\lVert\epsilon_\theta(x_t,t,z) - \epsilon\rVert_2^2$ for SDXL and ViT-DiT, always evaluated in the native VAE latent of the corresponding backbone. The reconstruction term is $\mathcal{L}_{\mathrm{rec}} = \mathbb{E}_t[(1{-}t)^2\lVert\hat x_0-x_0\rVert_2^2]$, which biases training toward cleaner timesteps; for the velocity parameterisation, $\hat x_0 = x_t - t v_\theta$. On top of this, we add a contrastive InfoNCE-style term $\mathcal{L}_{\mathrm{con}}$ on augmentation twins $(z,z')$, a VICReg term $\mathcal{L}_{\mathrm{vic}}$~\citep{bardes2022vicreg} with cross-rank all-gather ($\textsc{var\_w}{=}25$, $\textsc{cov\_w}{=}1$, target std $1$), and a frequency-domain loss $\mathcal{L}_{\mathrm{freq}}$ given by an $L_1$ distance between radially weighted log-amplitude FFT spectra of the decoded prediction and the target image. The FFT term is capped to at most two image pairs per batch and only evaluated for $t<0.7$ so that it improves high-frequency fidelity without destabilising high-noise updates.

\textbf{Loss-weight schedule}. The five $\lambda$s are the weightings of the loss components in Eq.~\eqref{eq:composite-loss}; they are not tuned independently and none were tuned per experiment. The \texttt{LossWeightScheduler} applies one fixed schedule to every ViT-FMDiT model: (i)~$\lambda_{\mathrm{diff}}$ ($0.5{\to}1.0$), $\lambda_{\mathrm{rec}}$ ($0.5{\to}0.1$), and $\lambda_{\mathrm{con}}$ ($0.3{\to}0.05$) are linearly interpolated between fixed endpoints; (ii)~the two regularisers are held at constant targets $\lambda_{\mathrm{vic}}{=}0.05$ and $\lambda_{\mathrm{freq}}{=}0.30$ but switched on with a delayed onset, i.e.\ exactly zero for the first epochs and then ramped linearly, so they cannot destabilise the encoder before it has settled. Only four numbers are free per weight (start, end, onset, ramp length); every intermediate value is determined. The remaining constants are not tuned: the VICReg sub-weights are the standard published values, and the frequency-loss sample cap is a memory bound. We use a two-phase encoder schedule: epochs $1$--$7$ update only the adaptors and bottleneck, then epoch $8$ unfreezes the upper $16$ ViT blocks, rebuilds the optimizer, and recomputes the cosine schedule in optimizer-step units. Training runs under PyTorch DDP on $4{\times}$H200 NVL with \texttt{find\_unused\_parameters=True} to tolerate the topology change; a per-rank skip handshake ensures that NaN-guarded steps are either taken or skipped synchronously across ranks.

\textbf{Loss-weight sensitivity}. We retrained ViT-FMDiT-$768$ under deliberate $\lambda$ perturbations with identical data (a fixed-split subset of $39{,}997$ micrographs covering $14$ alloys) and identical optimization settings, including a repeat of the reference configuration to measure the run-to-run noise floor, and scored each model on $4{,}000$ held-out micrographs (Tab.~\ref{tab:lambda-sensitivity}). Every variant converges monotonically (the largest epoch-to-epoch rise in validation loss is ${<}0.003$), and the final validation loss varies by only $0.6\%$ across the set. Removing the delayed onset, reducing $\lambda_{\mathrm{con}}$ six-fold, or changing four weights and both onsets at once shifts PSNR by at most $0.09$\,dB and grain disorientation by at most $0.21^\circ$, within a small multiple of the reference-versus-repeat noise floor; no configuration diverges or collapses. In archived runs spanning a $75$-fold range of $\lambda_{\mathrm{con}}$, a $10$-fold range of $\lambda_{\mathrm{diff}}$, and $\lambda_{\mathrm{vic}}\in[0,0.2]$, every configuration likewise converged monotonically.

\begin{table}[!ht]
  \caption{\textbf{Loss-weight sensitivity} of ViT-FMDiT-$768$ ($4{,}000$ held-out micrographs). The repeated reference measures the noise floor.}
  \label{tab:lambda-sensitivity}
  \centering
  \small
  \setlength{\tabcolsep}{4pt}
  \begin{tabular}{llccccc}
    \toprule
    Configuration & $\lambda$ change & Val.\ loss $\downarrow$ & PSNR $\uparrow$ & SSIM $\uparrow$ & LPIPS $\downarrow$ & Disor.\ ($^\circ$) $\downarrow$ \\
    \midrule
    reference             & committed schedule               & $0.4620$ & $12.47$ & $0.303$ & $0.592$ & $59.15$ \\
    reference (repeat)    & none (noise floor)               & $0.4621$ & $12.45$ & $0.298$ & $0.594$ & $58.99$ \\
    no delayed onset      & regularisers from epoch $0$      & $0.4595$ & $12.39$ & $0.285$ & $0.587$ & $58.94$ \\
    weak-reg + late onset & $4$ weights + both onsets        & $0.4603$ & $12.48$ & $0.299$ & $0.589$ & $59.05$ \\
    low $\lambda_{\mathrm{con}}$ & $\lambda_{\mathrm{con}}\times 1/6$ & $0.4604$ & $12.39$ & $0.291$ & $0.590$ & $59.08$ \\
    \midrule
    \textbf{spread (max$-$min)} & & $\mathbf{0.0026}$ & $\mathbf{0.09}$ & $\mathbf{0.018}$ & $\mathbf{0.007}$ & $\mathbf{0.21}$ \\
    \bottomrule
  \end{tabular}
\end{table}

\textbf{Bottleneck-capacity ablation ($d\in\{512,768,1024\}$)}. Bottleneck width is the most controlled ablation in this family because it leaves the decoder, encoder, conditioning path, and loss schedule unchanged. Under that constraint, held-out ViT-FMDiT PSNR improves monotonically with width, from roughly $14.0$ dB at epoch $2$ for $d{=}768$ to roughly $16.9$ dB at epoch $2$ for $d{=}1024$. This is consistent with the main-text choice of $d{=}512$ as a deliberately severe bottleneck for the latent-BO experiments rather than as the reconstruction-optimal point. Final-epoch reconstruction metrics for all three widths appear in Tab.~\ref{tab:reconstruction}.

\begin{table}[!ht]
  \caption{Decoder-specific hyperparameters on $4{\times}$H200. Batch size $=$ per-GPU $\times$ accum $\times$ \#GPUs.}
  \label{tab:hparams-decoder}
  \centering
  \footnotesize
  \setlength{\tabcolsep}{3pt}
  \begin{tabularx}{\linewidth}{@{}l X X X@{}}
    \toprule
    Hyperparameter & ViT-FMDiT & ViT-DiT & ViT-SDXL \\
    \midrule
    Backbone                          & SD3.5 MMDiT (${\sim}2.5$\,B) & DiT-XL/2 (${\sim}0.7$\,B) & SDXL UNet (${\sim}2.6$\,B) \\
    Backbone state                    & frozen        & \textbf{trainable}      & frozen \,(LoRA $r{=}16$) \\
    VAE                               & SD3.5 (16ch, $s{=}1.5305$) & sd-vae-ft-mse (4ch) & SD-XL VAE \\
    Bottleneck $d$ (head./abl.)       & $512/\{768,1024\}$ & $512/\{768,1024\}$ & $512$ \\
    LatentToTokens (\#q / depth)      & $16 / 5$      & $16$ spatial / --        & $77 / 3$ \\
    Conditioning dim                  & $4096$        & $1152$                   & $2048$ \\
    Pooled-projection dim             & $2048$        & ---                      & $1280$ \\
    Spatial injection blocks          & ---           & $\{0,7,14,21\}$          & --- \\
    Outer-proj init                   & zero          & Xavier ($0.10$)          & Xavier ($0.10$) \\
    Trainable interface               & ${\sim}0.6\%$ & $100\%$                  & ${\sim}0.4\%$ \\
    \midrule
    Per-GPU/accum/eff.\ batch         & $48/3/576$    & $144/1/576$              & $144/2/1152$ \\
    Train scheduler                   & FM-Euler      & DDPM (lin., $1000$)      & DDPM \\
    Inference scheduler ($N$ steps)   & FM-Euler ($50$) & DDIM ($250$)           & EulerDiscrete \\
    Timestep distribution             & logit-N, shift $3$ & uniform             & uniform \\
    CFG dropout                       & ---           & ---                      & $p_{\mathrm{drop}}{=}0.1$ \\
    \bottomrule
  \end{tabularx}
\end{table}

\begin{figure}[!ht]
  \centering
  \includegraphics[width=\linewidth]{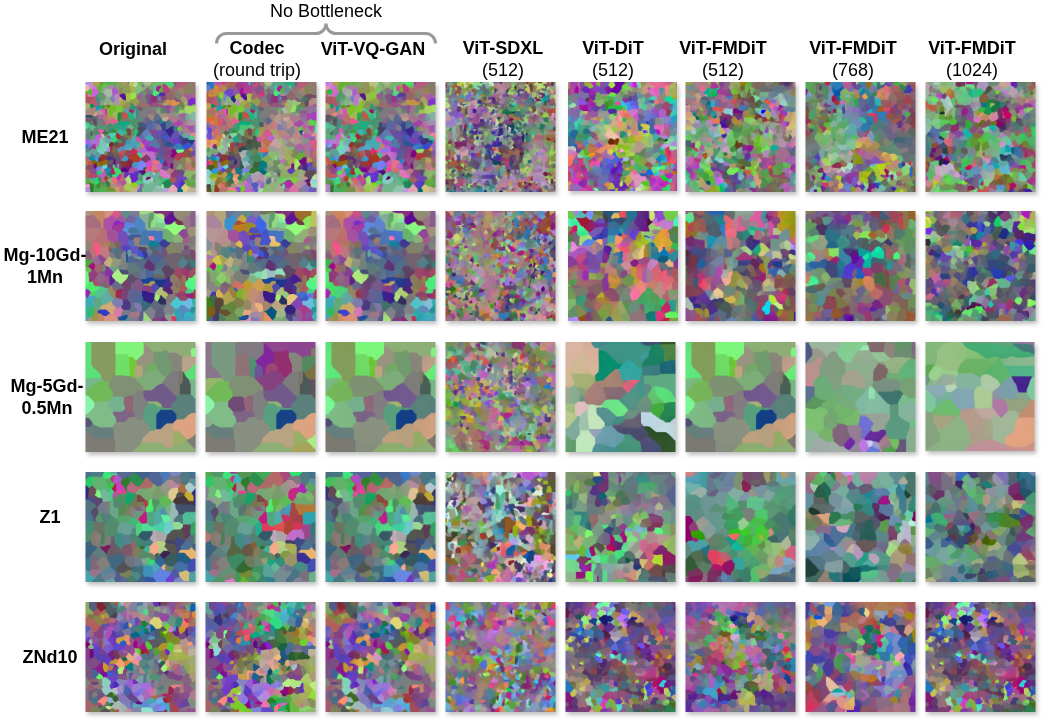}
  \caption{\textbf{Extended reconstruction gallery on five additional alloy classes} (rows, in order: \texttt{ME21}, \texttt{ZX10}, \texttt{Mg-5Gd-0.5Mn}, \texttt{Z1}, \texttt{ZNd10}); columns follow Fig.~\ref{fig:reconstruction}.}
  \label{fig:reconstruction-extended}
\end{figure}

\section{Orientation codec --- extended theory and round-trip diagnostics}
\label{app:codec-extended}

This section expands the codec analysis from Sec.~\ref{sec:codec} and makes explicit the geometric choices that let a generic image decoder serve as a microstructure prior. The thread is simple: every step is chosen to preserve local orientation structure while keeping the RGB image numerically stable for downstream generative models.

\textbf{Quaternion parameterisation and HCP symmetry}. We first convert each per-grain orientation from Bunge--Euler angles ($ZXZ$) to a unit quaternion $q=(q_x,q_y,q_z,q_w)\in S^3$~\citep{zhou2019}. Because quaternions form a double cover ($q\equiv -q$), we fix the sign by enforcing the hemisphere convention $q_w\ge 0$. The relevant crystal symmetry is the HCP point group $D_6$ with $|\mathcal{S}_{\mathrm{HCP}}|=12$, generated by the six $c$-axis rotations $R_z(k\pi/3)$, $k=0,\ldots,5$, and the six compositions $R_z(k\pi/3)\cdot R_x(\pi)$. In quaternion coordinates $R_z(\theta)=(\sin(\theta/2)\,\hat z,\cos(\theta/2))$ under the standard Hamilton product. Two orientations are therefore physically identical whenever they differ by a symmetry element $s_k \in \mathcal{S}_{\mathrm{HCP}}$.

\textbf{Step 1 --- continuous unfolding (anchored BFS)}. Let $\mathcal{N}=(\mathcal{V},\mathcal{E})$ denote the grain adjacency graph from \dreamthreed{}, with $|\mathcal{V}|=G$ grains and $\mathcal{E}$ the set of $4$-connected boundary pairs. We choose the BFS root deterministically as the largest grain,
\begin{equation}
  g^\star = \arg\max_{g \in \mathcal{V}}\; \bigl|\{(i,j) : \mathcal{G}[i,j] = g\}\bigr|,
  \label{eq:bfs-root}
\end{equation}
and align its quaternion to the class anchor (Step 2) by $q_{g^\star} \leftarrow \arg\max_{s,\sigma}\, \sigma\,\langle s\cdot q_{g^\star},\, \bar{q}_c\rangle$ over $s\in\mathcal{S}_{\mathrm{HCP}}$ and $\sigma\in\{\pm 1\}$, where $\langle\cdot,\cdot\rangle$ is the $\R^4$ inner product. Using the largest grain as the seed shortens the average BFS path length, and aligning that single seed to $\bar q_c$ keeps the unfolded tree on a common hemisphere so that identical orientations map to consistent colours across the class. Each remaining grain then inherits, from its already visited neighbour $g_p$, the symmetry equivalent and sign with maximal inner product:
\begin{equation}
  q_{g_c} \;\leftarrow\; \arg\max_{s,\sigma}\; \sigma\, \langle s \cdot q_{g_c},\; q_{g_p} \rangle.
  \label{eq:bfs-step}
\end{equation}
Each edge requires $24$ dot products, so the encode remains $\mathcal{O}(|\mathcal{E}|)$ in the boundary count. After this pass, quaternion distance between adjacent grains tracks their physical misorientation directly, without the artificial jumps induced by per-grain folding to a fundamental zone.

\textbf{Step 2 --- class anchor via Markley averaging}. For each class $c$, the anchor $\bar q_c$ is the $L_2$-optimal mean of $M{=}50$ representative file-level mean quaternions $\{m_j\}_{j=1}^M$. Following \citet{markley2007}, it is the leading eigenvector of
\begin{equation}
  A = \frac{1}{M}\sum_{j=1}^{M} m_j\, m_j^{\top} \;\in\; \R^{4\times 4}.
  \label{eq:markley}
\end{equation}
The point of using a class-level rather than per-sample anchor is that generated images arrive without per-sample metadata. The cost is a mild long-range colour drift induced by within-class spread, but at $M{=}50$ that drift is still comfortably inside the capacity of the encoder bottleneck.

\textbf{Steps 3--4 --- frame centring and stereographic projection}. Once the anchor is factored out via $q_g \leftarrow \bar q_c^{-1}\cdot q_g$, the distribution concentrates near identity ($q_w\approx 1$, $\norm{q_{xyz}}\ll 1$). Enforcing $q_w\ge 0$, we project to
\begin{equation}
  S \;=\; \frac{(q_x,\, q_y,\, q_z)}{1 + q_w} \;\in\; [-1, 1]^3,
  \label{eq:stereo}
\end{equation}
which is $C^\infty$ on the open hemisphere $q_w>0$. The inverse map is closed form,
\begin{equation}
  q_w \;=\; \frac{1 - \norm{S}^2}{1 + \norm{S}^2}, \qquad
  (q_x, q_y, q_z) \;=\; \frac{2\, S}{1 + \norm{S}^2},
  \label{eq:istereo}
\end{equation}
and uses only rational arithmetic. This is the crucial numerical simplification: there is no square root, so decoder noise cannot drive the inverse into NaNs. As $\norm{S}\to\infty$, $q_w$ approaches $-1$ smoothly rather than catastrophically.

\textbf{Step 5 --- quantisation error bound}. For a centred unit quaternion near identity, let $S=q_{xyz}/(1+q_w)$ and perturb it by $\delta S$. To first order, $\delta q_{xyz}\approx (1+q_w)\,\delta S$ while $\delta q_w$ is second order. The induced angular error is therefore $\delta\theta = 2\arcsin(\norm{\delta q_{xyz}}) \approx 2(1+q_w)\norm{\delta S} \approx 2\norm{\delta S}$ near identity. With quantisation step $\Delta = 2/(2^b{-}1)$, this yields the worst-case bound $\delta\theta \lesssim 0.78^\circ$ for $b{=}8$ and $0.004^\circ$ for $b{=}16$. The measured PNG mean in Tab.~\ref{tab:codec-accuracy} is $0.6^\circ$, in line with that estimate.

\begin{table}[!ht]
  \caption{Codec round-trip accuracy on three Mg-alloy classes ($300{\times}300$ images, $257$--$2{,}044$ grains/sample). Boundary F1 and grain-size Pearson correlation are unchanged between TIFF and PNG; only quantisation noise differs.}
  \label{tab:codec-accuracy}
  \centering
  \small
  \begin{tabular}{lcccc}
    \toprule
    Class & Boundary F1 & Grain-size $r$ & PNG-vs-TIFF mean ($^\circ$) & PNG-vs-TIFF max ($^\circ$) \\
    \midrule
    \texttt{ME21\_extruded}    & $0.9999$ & $0.9997$ & $0.61$ & $1.35$ \\
    \texttt{Mg{-}10Gd\_extruded} & $1.0000$ & $0.9993$ & $0.69$ & $1.30$ \\
    \texttt{AZ31\_extruded\_HT} & $0.9999$ & $0.9991$ & $0.61$ & $1.37$ \\
    \bottomrule
  \end{tabular}
\end{table}

\textbf{Why these choices}. The design logic is now clear. The rational inverse in \eqref{eq:istereo} replaces the unstable recovery $\sqrt{1-q_x^2-q_y^2-q_z^2}$, which fails as soon as decoder noise makes the radicand negative. Continuous unfolding preserves the proportionality between RGB distance and true misorientation, keeping the encoded image in the low-frequency regime that pretrained image backbones model well. The class-level Markley anchor is not mathematically ideal, but it is the only option compatible with generated images that lack per-sample metadata. Finally, we use $b{=}8$ inside \Co{} because pretrained ViT and diffusion backbones expect 8-bit RGB, and the resulting $\sim 0.6^\circ$ floor remains below both typical EBSD angular resolution and the Read--Shockley low-angle threshold.

\section{DAMASK and the HCP phenopower-law constitutive model ($\mathbf{F}_p$, $\dot\gamma^\alpha$, $\xi^\alpha$)}
\label{app:damask}

This section records the constitutive model and post-processing pipeline actually used by the oracle. The main text only needs the fact that \damask{} returns a stress--strain response; here we spell out the kinematics, hardening law, and property extraction so the simulation layer is unambiguous.

\textbf{Kinematics and resolved shear stress}. We use the standard multiplicative split $\mathbf{F}=\mathbf{F}_e\mathbf{F}_p$, which separates elastic lattice stretch from plastic shear~\citep{Wang2021}. Plastic flow is written as the sum over all active slip and twin systems,
$\mathbf{L}_p = \sum_\alpha \dot\gamma^\alpha\, \mathbf{s}^\alpha \otimes \mathbf{m}^\alpha$,
with $\mathbf{s}^\alpha$ and $\mathbf{m}^\alpha$ the unit slip or twin direction and plane normal. The resolved shear stress on system $\alpha$ is then $\tau^\alpha = \boldsymbol{\sigma}\!:\!(\mathbf{s}^\alpha\otimes\mathbf{m}^\alpha)$.

\textbf{Power-law flow rule}. Shear rates follow a phenomenological power law, separately on slip ($sl$) and twin ($tw$) systems:
\begin{equation}
  \dot\gamma^\alpha_{sl} = \dot\gamma_0^{sl}\,\Bigl|\tfrac{\tau^\alpha}{\xi^\alpha_{sl}}\Bigr|^{n_{sl}}\!\mathrm{sign}(\tau^\alpha),
  \qquad
  \dot\gamma^\alpha_{tw} = \dot\gamma_0^{tw}\,\Bigl(\tfrac{\tau^\alpha}{\xi^\alpha_{tw}}\Bigr)^{n_{tw}},\;\;\tau^\alpha > 0,
  \label{eq:phenopower-flow}
\end{equation}
with reference rates $\dot\gamma_0^{sl/tw}$, rate-sensitivity exponents $n_{sl/tw}$, and critical resolved shear stresses $\xi^\alpha_{sl/tw}$. Twinning contributes only for $\tau^\alpha>0$, reflecting the polarity of twin shear in HCP crystals~\citep{ActaMg2014}.

\textbf{Hardening evolution}. The CRSS evolves through self- and latent-hardening interactions,
$\dot\xi^\alpha = \sum_\beta h^{\alpha\beta}\,|\dot\gamma^\beta|$,
where $h^{\alpha\beta}$ is the system-interaction matrix. Slip systems use saturation-type hardening with explicit slip--twin coupling,
\begin{equation}
  \dot\xi^\alpha_{sl} = h_0^{sl}\,\bigl(1 - \xi^\alpha_{sl}/\xi^\infty_{sl}\bigr)^{a_{sl}}\!\sum_\beta h^{\alpha\beta}_{sl\text{-}sl}\,|\dot\gamma^\beta_{sl}|
     + f^{sl\text{-}tw}_{sat}\!\sum_\beta h^{\alpha\beta}_{sl\text{-}tw}\,|\dot\gamma^\beta_{tw}|,
  \label{eq:phenopower-slip-hardening}
\end{equation}
with initial slip-hardening modulus $h_0^{sl}$, saturation CRSS $\xi^\infty_{sl}$, hardening exponent $a_{sl}$, and slip--twin coupling factor $f^{sl\text{-}tw}_{sat}$. Twin systems harden symmetrically through twin--twin and twin--slip channels,
\begin{equation}
  \dot\xi^\alpha_{tw} = h_0^{tw\text{-}tw}\!\sum_\beta h^{\alpha\beta}_{tw\text{-}tw}\,|\dot\gamma^\beta_{tw}|
     + h_0^{tw\text{-}sl}\!\sum_\beta h^{\alpha\beta}_{tw\text{-}sl}\,|\dot\gamma^\beta_{sl}|.
  \label{eq:phenopower-twin-hardening}
\end{equation}
The model is intentionally phenomenological: hardening is encoded through these scalar moduli rather than through dislocation-density or twin-volume-fraction state variables~\citep{Wang2021}. That choice keeps the oracle tractable while preserving the slip--twin asymmetry that matters most in HCP magnesium~\citep{ActaMg2014}. Although damage-augmented variants exist~\citep{FeCrAl2021}, we do not use them here. Instead, every run uses the AZ31-calibrated parameter file distributed with \damask{} (\texttt{AZ31\_Phenopower.yaml}), with basal, prismatic, and pyramidal-$\langle a\rangle$/-$\langle c{+}a\rangle$ slip plus tensile twinning, under isothermal uniaxial tension along the extrusion direction and traction-free transverse faces.

\textbf{Hollomon fit and property extraction}. Post-processing is straightforward but consistent across all runs. We read the HDF5 output, compute volume-averaged Cauchy stress and logarithmic strain along the loading axis at every increment, identify the elastic--plastic transition by the $0.2\%$ offset rule, and fit the plastic branch in log--log coordinates with the Hollomon law $\sigma = K\varepsilon_p^n$. The yield stress $\sigy$ is the offset intercept, $\sigu$ the maximum stress, and $(K,n)$ the fitted Hollomon parameters. Runs that do not converge, contain NaN stresses, or produce fewer than $50$ converged increments are discarded and logged.

\begin{table}[!ht]
  \caption{\damask{} simulation parameters used for all property evaluations. The same \texttt{load.yaml} and \texttt{AZ31\_Phenopower.yaml} are used across every cell of E2; only the input \texttt{.dream3d} (i.e., the decoded microstructure) varies.}
  \label{tab:hparams-sim}
  \centering
  \small
  \begin{tabular}{ll}
    \toprule
    Parameter & Value \\
    \midrule
    Loading mode                  & uniaxial tension along ED \\
    Total time $t$ (s)            & $250$ \\
    Increments $N$                & $1250$ \\
    Strain rate $\dot{\varepsilon}$ (s$^{-1}$) & $1.0 \times 10^{-3}$ \\
    Output frequency $f_{\mathrm{out}}$ & every increment \\
    Constitutive law              & phenopowerlaw (HCP) \\
    Slip families                 & basal, prismatic, pyr.\ $\langle a\rangle$, pyr.\ $\langle c{+}a\rangle$ \\
    Twinning                      & tensile twin only \\
    Phase parameter file          & \texttt{AZ31\_Phenopower.yaml} \\
    Failure threshold             & ${<}50$ converged increments $\Rightarrow$ drop \\
    \bottomrule
  \end{tabular}
\end{table}

\section{Baseline optimizer implementations and hyperparameters}
\label{app:baselines}

This appendix gathers the implementation detail for the optimizer benchmark that would be distracting in the main text: the pipeline-level adaptations applied uniformly across methods, the concrete baseline settings, the compact design comparison in Tab.~\ref{tab:opt-comparison}, and the full hyperparameter table in Tab.~\ref{tab:hparams-opt}. All values are those committed in \texttt{config/az31\_*.yaml} and \texttt{copilot/optimizers/} of the released code; no per-decoder, per-objective, or per-seed retuning was introduced after the sweep design was fixed.

\subsection{Pipeline-level adaptations applied to the latent-BO optimizers}

\textbf{Surrogate masking of failed evaluations.}
The decode$\to$codec$\to$\damask{} pipeline can reject a candidate for several reasons: the decoded image may collapse to blank or speckle structure, it may fail the chromatic-spread gate, it may produce fewer than $g_{\min}{=}40$ grains, or the simulator may fail to converge. Following standard latent-BO practice~\citep{eriksson2019,papenmeier2022}, we log such cases for bookkeeping but exclude them from the regression surrogate in \dante{}, \turbo{}, \baxus{}, and \meridian{}. A constant penalty would pollute posterior variance without adding useful structure and would do so symmetrically across baselines. \meridian{} is the only method that still retains these failures explicitly through its feasibility classifier $g_\psi$.

\textbf{Trust-region restart cap.}
The second global intervention is the restart cap. Under the published \turbo{} rule $\tau_{\mathrm{fail}}=\lceil d/q \cdot \alpha \rceil$, the failure tolerance becomes $192$ for \turbo{} and \meridian{} at $(d,q,\alpha)=(512,4,1.5)$, and $128$ for \baxus{} at its initial subspace dimension $d_0=128$. In a $200$-evaluation regime that effectively disables restarts and turns all three methods into one-shot local searches. We therefore impose a shared guardrail $\tau_{\mathrm{fail}}^{\max}=8$ on the three trust-region methods while leaving the published $\alpha$ unchanged. The point is not to strengthen any one method, but to make the documented restart logic reachable at all. \dante{} has no trust-region mechanism and is unchanged.

\subsection{\texorpdfstring{\dante{}~\citep{wei2025}}{DANTE}}

For \dante{}, we use the authors' Neural Tree Explorer (NTE): an iterative root-and-cloud DUCB tree search paired with a deep MLP surrogate. The surrogate has hidden widths $[1024,512,256]$, dropout $p=0.1$, and is trained for $200$ epochs with AdamW (lr $10^{-3}$, weight decay $10^{-4}$, batch size $32$, validation split $0.1$, early-stopping patience $20$). Each round expands $200$ leaves with branching factor $16$ and depth-decayed Gaussian leaves using $\sigma_{\mathrm{init}}=0.05$, $\sigma_{\mathrm{decay}}=0.995$, DUCB constant $c_0=0.1$, and smoothing $\rho=0.5$. Since \dante{} has no native restart mechanism, we leave its visit-accumulation behaviour untouched.

\subsection{\texorpdfstring{\turbo{}~\citep{eriksson2019}}{TuRBO}}

\turbo{} is run as a single trust-region GP method with an ARD-Mat\'ern-$5/2$ kernel and Thompson sampling over $5{,}000$ Sobol candidates. The success/failure cadence is the standard one, $\tau_{\mathrm{succ}}{=}3$ and $\tau_{\mathrm{fail}}{=}\lceil d/q \cdot \alpha\rceil$ with $\alpha=1.5$, but with the shared cap $\tau_{\mathrm{fail}}^{\max}{=}8$ described above. Trust-region bounds are $L\in[0.005,1.6]$ with $L_{\mathrm{init}}=0.8$, and collapse triggers a Sobol cold start in $\Z$. We also activate a plateau-restart guardrail (B1): after $K_{\mathrm{plat}}{=}5$ non-improving batches, the method is restarted even if the trust region has not yet formally collapsed. Without this extra trigger, the published shrink schedule would require about seven halvings to reach $L_{\min}$, which does not fit inside our $40$-iteration budget.

\subsection{\texorpdfstring{\baxus{}~\citep{papenmeier2022}}{BAxUS}}

\baxus{} uses the same acquisition and trust-region cadence as \turbo{}, but fits its GP in a low-dimensional embedded subspace. We begin at the authors' default target dimension $d_0=10$ and replace the usual random sparse $\pm 1$ embedding with a PCA-of-seeds basis (B2): the projector columns are the leading PCA directions of the class-conditional latent pool $\mathcal P$, truncated to $k_{\mathrm{eff}}=\min(d_0,\mathrm{rank}(\mathcal P))$. This keeps lifted proposals aligned with the encoded seed support. We also enable reject feedback to the GP (\texttt{wants\_reject\_feedback = True}) so that soft-floor codec failures are treated as informative evidence about bad directions rather than discarded.

\subsection{Additional baselines: CMA-ES, DDOM, SEIKO, and DDPO}
\label{app:extra-baselines}

These four baselines share the decoder, latent box, $100$-simulation seed cache, batch size $q{=}4$, $160$-evaluation budget, and five seeds with the methods above, and are run on ViT-FMDiT-$768$ and -$1024$. CMA-ES and DDOM are direct optimizer-level comparisons to \meridian{}; SEIKO and DDPO modify the decoder itself and therefore serve as framework-level comparisons to \Co{}.
\textbf{CMA-ES}~\citep{hansen2001} maintains a Gaussian search distribution over $\Z$, evaluates samples, and adapts its mean and covariance toward high-scoring candidates. It has no surrogate of the design space, so everything it learns is paid for in simulator calls, whereas \meridian{} fits a predictive model to every past simulation before proposing.
\textbf{DDOM}~\citep{krishnamoorthy2023ddom} learns a generative model over designs conditioned on objective value and samples toward high values; adapted from its original offline setting to our fixed-budget loop, it is retrained each round to lean toward the best result seen so far. It proposes plausible candidates but checks neither whether they will run in the simulator nor localises search around the incumbent, the two capabilities \meridian{} adds through $g_\psi$ and the trust region.
\textbf{SEIKO}~\citep{uehara2024seiko} repeatedly fine-tunes the decoder toward high reward while keeping it close to its previous version for stability. It improves the generator's \emph{average} tendency to produce good designs rather than exploiting the single best design found so far, which limits it under our budget.
\textbf{DDPO}~\citep{black2024ddpo} treats denoising as a sequential decision process and fine-tunes the decoder by policy gradient on the reward. It is data-hungry: its reference configurations use $25{,}600$--$57{,}600$ reward queries, $160$--$360\times$ our budget.

\begin{table}[!ht]
  \caption{Key design distinctions among the four latent-optimizers benchmarked on every decoder. Mechanisms unique to \meridian{} are highlighted in bold. The additional baselines of App.~\ref{app:extra-baselines} have no feasibility model, trust region, or manifold prior.}
  \label{tab:opt-comparison}
  \centering
  \small
  \setlength{\tabcolsep}{4pt}
  \begin{tabularx}{\linewidth}{l XXXX}
    \toprule
    Design axis & \dante{}~\citep{wei2025} & \turbo{}~\citep{eriksson2019} & \baxus{}~\citep{papenmeier2022} & \meridian{} (ours) \\
    \midrule
    Surrogate
      & MLP point estimate
      & ARD-Mat\'ern-$5/2$ GP
      & ARD-Mat\'ern-$5/2$ GP in subspace
      & \textbf{deep-kernel GP} \\
    Target-aware signal
      & scalar-$Y$ only
      & scalar-$Y$ only
      & scalar-$Y$ only
      & \textbf{property-head EI for $V2$} \\
    Feasibility handling
      & none
      & none
      & reject feedback to GP
      & \textbf{shared-trunk classifier} $g_\psi$ \\
    Proposal geometry
      & global tree expansion
      & isotropic trust region
      & low-dim subspace trust region
      & \textbf{anisotropic trust region} \\
    Manifold prior
      & none
      & none
      & PCA-of-seeds embedding
      & \textbf{active subspace + adaptive shell} \\
    Batch strategy
      & top-$k$ predicted
      & max-posterior sampling
      & max-posterior sampling
      & \textbf{quality-weighted DPP} \\
    Restart rule
      & none
      & Sobol restart + plateau cap
      & subspace doubling
      & \textbf{class-pool or mini-MCTS restart} \\
    Final-phase behavior
      & monotone
      & monotone
      & monotone
      & \textbf{explicit polish phase} \\
    \bottomrule
  \end{tabularx}
\end{table}

\subsection{\meridian{} (ours)}
\label{app:meridian-details}

\begin{table*}[t]
  \caption{Optimizer hyperparameters for the four latent-BO optimizers; all share batch size $q=4$, latent dimension $d=512$, search box $\Z=[-3,3]^{512}$, $T=40$ outer iterations, and seed-cache warm start $n_0=100$.}
  \label{tab:hparams-opt}
  \centering
  \scriptsize
  \setlength{\tabcolsep}{3pt}
  \renewcommand{\arraystretch}{1.04}
  \begin{minipage}[t]{0.48\textwidth}
    \centering
    \textbf{\dante{}}\\[1pt]
    \begin{tabular}{@{}p{0.38\linewidth}p{0.56\linewidth}@{}}
      \toprule
      Model & MLP $[1024,512,256]$ \\
      Acquisition & DUCB \\
      Candidates & $200$ leaves \\
      Geometry & branching $16$ tree \\
      Training & AdamW, $200$ epochs \\
      lr / wd & $10^{-3}$ / $10^{-4}$ \\
      Regularisation & dropout $0.1$ \\
      Exploration & $\sigma_{\mathrm{init}}=0.05$ \\
      Decay & $\sigma_{\mathrm{decay}}=0.995$ \\
      Score & $c_0=0.1$, $\rho=0.5$ \\
      Restart & none \\
      \\
      \bottomrule
    \end{tabular}
  \end{minipage}\hfill
  \begin{minipage}[t]{0.48\textwidth}
    \centering
    \textbf{\turbo{}}\\[1pt]
    \begin{tabular}{@{}p{0.38\linewidth}p{0.56\linewidth}@{}}
      \toprule
      Surrogate & ARD-Mat\'ern-$5/2$ GP \\
      Acquisition & Thompson \\
      Candidates & $5{,}000$ Sobol \\
      Geometry & isotropic trust region \\
      Trust region & $L_{\mathrm{init}}=0.8$, $L_{\min}=0.005$, $L_{\max}=1.6$ \\
      Success rule & $\tau_{\mathrm{succ}}=3$ \\
      Failure rule & $\alpha=1.5$, $\tau_{\mathrm{fail}}^{\max}=8$ \\
      Plateau cap & $K_{\mathrm{plat}}=5$ \\
      Warm start & Sobol cache \\
      Restart & Sobol cold-start \\
      Polish & --- \\
      \bottomrule
    \end{tabular}
  \end{minipage}

  \vspace{4pt}

  \begin{minipage}[t]{0.48\textwidth}
    \centering
    \textbf{\baxus{}}\\[1pt]
    \begin{tabular}{@{}p{0.38\linewidth}p{0.56\linewidth}@{}}
      \toprule
      Surrogate & ARD-Mat\'ern-$5/2$ GP \\
      Acquisition & Thompson \\
      Candidates & $5{,}000$ Sobol \\
      Geometry & subspace trust region \\
      Subspace & PCA-of-seeds, $d_0=10$ \\
      Warm start & PCA of $\mathcal{P}$ \\
      Trust region & $L_{\mathrm{init}}=0.8$, $L_{\min}=0.005$, $L_{\max}=1.6$ \\
      Success rule & $\tau_{\mathrm{succ}}=3$ \\
      Failure rule & $\alpha=1.5$, $\tau_{\mathrm{fail}}^{\max}=8$ \\
      Reject feedback & on \\
      Restart & subspace doubling \\
      Polish & --- \\
      \\
      \bottomrule
    \end{tabular}
  \end{minipage}\hfill
  \begin{minipage}[t]{0.48\textwidth}
    \centering
    \textbf{\meridian{}}\\[1pt]
    \begin{tabular}{@{}p{0.38\linewidth}p{0.56\linewidth}@{}}
      \toprule
      Trunk & MLP $[512,128]\to 16$ \\
      Heads & feasibility + property ($V2$) \\
      Acquisition & qLogNEI + prop. MC-EI \\
      Candidates & $4{,}096$ Sobol \\
      Geometry & anisotropic trust region \\
      Trust region & $L_{\mathrm{init}}=0.6$, $L_{\min}=0.05$, $L_{\max}=1.0$ \\
      Success rule & $\tau_{\mathrm{succ}}=3$ \\
      Failure rule & $\alpha=0.5$, $\tau_{\mathrm{fail}}^{\max}=8$ \\
      Shell / exempt & $k_\sigma=2$, $\rho_{\mathrm{exempt}}=0.15$ \\
      Diffuse / sparse & $\rho_{\mathrm{diff}}=0.20$, $20/512$ axes \\
      Batch / DPP & top-$256$, $w_z=0.05$ \\
      Restart / polish & pool or mini-MCTS; $T_{\mathrm{pol}}=35$ \\
      Plateau / cap & $K_{\mathrm{plat}}=5$, $r_{\max}=3$ \\
      \bottomrule
    \end{tabular}
  \end{minipage}
\end{table*}

This subsection records the implementation choices that are intentionally omitted from Sec.~\ref{sec:meridian}. The emphasis here is not on re-deriving the main algorithm, but on documenting the concrete engineering choices that make the method stable in the expensive-oracle regime. All numerical settings appear in Tab.~\ref{tab:hparams-opt} and the committed values in \texttt{config/az31\_*\_meridian\_v2.yaml}. The surrogate stack is a deep-kernel model in which a two-hidden-layer MLP $\phi_\theta : \R^{512} \to \R^{16}$ (widths $[512,128]$, LayerNorm + GELU) is trained jointly with the feasibility classifier $g_\psi$ on \emph{all} observations, preserving the decode/codec/simulator failure signal, and the exact GP is then refit only on the feasible subset. For target-driven $V2$, the same feature trunk also feeds a heteroscedastic linear property head $h_p : \R^{16} \to \R^{2\cdot 5}$ that predicts $(\mu,\log\sigma^2)$ for $(\sigma_y,\sigma_u,n,K,n_{\mathrm{grains}})$ in standardised units. The acquisition used in practice is therefore
\begin{equation}
  \alpha(z) =
  \begin{cases}
    \alpha_{\mathrm{GP}}(z), & V1,\\
    \tfrac{1}{2}\bigl[\tilde\alpha_{\mathrm{GP}}(z) + \tilde\alpha_{\mathrm{prop}}(z)\bigr], & V2,
  \end{cases}
  \qquad
  \alpha_{\mathrm{GP}}(z) = \mathrm{qLogNEI}(z;\mu,\sigma)\, g_\psi(z),
\end{equation}
where $\tilde\alpha_{\mathrm{GP}}$ and $\tilde\alpha_{\mathrm{prop}}$ are $z$-standardised scores and $\tilde\alpha_{\mathrm{prop}}$ is estimated by Monte-Carlo EI with $64$ samples. The property-head linear layer is reset every $K_{\mathrm{reset}}{=}10$ refits to avoid variance collapse on a saturated cache, while the shared feature trunk is warm-started throughout.

Proposal generation is tied explicitly to the latent manifold learned by the decoder. After the first two rounds, search anisotropy is read off from the active-subspace matrix
\begin{equation}
  \hat C = \frac{1}{N}\sum_{i=1}^{N} \nabla \hat\mu(z_i)\, \nabla \hat\mu(z_i)^\top,
\end{equation}
whose leading eigenspace is truncated at $95\%$ cumulative energy and reprojected to ambient axis weights $w\in\R^{512}$; during cold start, that role is played instead by PCA on the class-conditional latent pool $\mathcal P$. Around the shell-projected centroid of the top-$K{=}5$ feasible incumbents, \meridian{} draws a Sobol cloud of $N{=}4096$ candidates, perturbs only a sparse mask of axes with $\Pr[M_{ij}{=}1]=20/d$, and replaces a fraction $\rho_{\mathrm{diff}}{=}0.20$ with isotropic Gaussian moves at scale $0.5L_t\bar w$ so that diffuse high-value pockets are not ruled out a priori. Most proposals are then clipped to the adaptive shell
\begin{equation}
  \|z\| \in [\mu_{\|X\|} \pm k_\sigma \, \sigma_{\|X\|}],
  \qquad k_\sigma = 2,
\end{equation}
estimated from the feasible samples, while a small exempt fraction $\rho_{\mathrm{exempt}}{=}0.15$ is left off-shell in case the support genuinely needs to widen. The trust-region edge evolves in the range $L_t\in[0.05,1.0]$ with $L_{\mathrm{init}}{=}0.6$ under the same success/failure cadence used by \turbo{}/\baxus{}.

Batch construction and restart are likewise tuned for the expensive-oracle regime rather than for formal neatness. From the top-$256$ candidates by acquisition value, a greedy DPP selects the final batch with kernel
\begin{equation}
  L_{ij} = \alpha_i\alpha_j\Bigl[(1-w_z) \, k_{\mathrm{GP}}\bigl(\phi(z_i),\phi(z_j)\bigr) + w_z \, k_{\mathrm{RBF}}(z_i,z_j)\Bigr],
  \qquad w_z = 0.05,
\end{equation}
so diversity is enforced mainly in feature space without giving up direct latent-space separation. For $V2$, one slot is overwritten by a short property-gradient seed and one by a jittered incumbent-polish seed. Past $T_{\mathrm{pol}}{=}35$ of $T{=}40$ rounds, the diffuse fraction, shell exemption, and latent-RBF blend are all set to zero and $L_t$ is capped at $0.1$, yielding a deliberately local polishing phase. A restart is triggered either by trust-region collapse ($L_t<L_{\min}$) or by a plateau of $K_{\mathrm{plat}}{=}5$ rounds; the next anchor is drawn from the class pool or from a small surrogate-side mini-MCTS, the property heads are reset, and termination occurs after $r_{\max}{=}3$ unsuccessful restarts. \refstepcounter{algorithm}\label{alg:meridian-full}\noindent\textbf{Algorithm~\thealgorithm.} This narrative description is the full supplementary specification referenced from the main text.

We use each baseline at the configuration recommended by its authors, with the two pipeline-level adaptations described above (failure masking; restart cap) applied uniformly across the four latent-BO optimizers, plus the two manifold-awareness adjustments to \baxus{} (PCA-of-seeds embedding, soft-floor reject feedback) detailed in the \baxus{} subsection above. The committed hyperparameter values are listed in Tab.~\ref{tab:hparams-opt}.

\section{Reproducibility checklist mapping}
\label{app:repro}

\begin{table}[H]
  \caption{Reproducibility checklist items to the section, file, or asset that satisfies them.}
  \label{tab:repro}
  \centering
  \small
  \setlength{\tabcolsep}{4pt}
  \begin{tabularx}{\linewidth}{@{}l l X@{}}
    \toprule
    Checklist item & Where addressed & Asset \\
    \midrule
    Code released                  & Secs.~\ref{sec:intro},~\ref{sec:conclusion} & encoder--decoder: \urlEncDec{}; orientation codec: \urlCodec{}; \meridian{} and \damask{} harness: \urlMeridian{} \\
    Decoder hyperparameters        & App.~\ref{app:enc-dec}           & Tabs.~\ref{tab:hparams-decoder},~\ref{tab:lambda-sensitivity} \\
    Simulator hyperparameters      & App.~\ref{app:damask}            & Tab.~\ref{tab:hparams-sim} \\
    Optimizer hyperparameters      & App.~\ref{app:baselines}         & Tab.~\ref{tab:hparams-opt} \\
    Sensitivity / ablations        & App.~\ref{app:ablations}         & Tabs.~\ref{tab:dpp-topk},~\ref{tab:meridian-sensitivity} \\
    Compute disclosed              & Sec.~\ref{sec:e1}; Sec.~\ref{sec:e2}; App.~\ref{app:extended-results}       & --- \\
    Limitations stated             & Sec.~\ref{sec:conclusion}        & --- \\
    Training Dataset               & Sec.~\ref{sec:e1}                & \urlData{} \\
    \bottomrule
  \end{tabularx}
\end{table}


\end{document}